\pdfoutput=1
\documentclass[letterpaper]{article} 
\usepackage{aaai25}  
\usepackage{times}  
\usepackage{helvet}  
\usepackage{courier}  
\usepackage[hyphens]{url}  
\usepackage{graphicx} 
\usepackage{natbib}  
\usepackage{caption} 
\usepackage{algorithm}
\usepackage{algorithmic}

\usepackage{booktabs,multirow,adjustbox,diagbox,threeparttable}
\usepackage{amssymb}
\usepackage{xspace}
\usepackage{makecell}

\usepackage{newfloat}
\usepackage{listings}
\DeclareCaptionStyle{ruled}{labelfont=normalfont,labelsep=colon,strut=off} 
\floatstyle{ruled}
\newfloat{listing}{tb}{lst}{}
\floatname{listing}{Listing}
\def\eg{\emph{e.g.}}

\newcommand{\method}{SAIL\xspace}

\newcommand{\vs}{\textit{vs.}\xspace}

\title{SAIL: Sample-Centric In-Context Learning for Document Information Extraction}
\author{
    Jinyu Zhang\textsuperscript{\rm 1}\equalcontrib,
    Zhiyuan You\textsuperscript{\rm 2}\equalcontrib,
    Jize Wang\textsuperscript{\rm 1},
    Xinyi Le\textsuperscript{\rm 1}\thanks{Corresponding author}\\
}
\affiliations{
    \textsuperscript{\rm 1}Shanghai Jiao Tong University\quad
    \textsuperscript{\rm 2}The Chinese University of Hong Kong\\

    \{zhang\_jinyu, jizewang2000, lexinyi\}@sjtu.edu.cn \quad zhiyuanyou@foxmail.com
}

\usepackage{bibentry}

\begin{document}

\maketitle

\begin{abstract}\label{sec:abs}
Document Information Extraction (DIE) aims to extract structured information from Visually Rich Documents (VRDs). 
Previous full-training approaches have demonstrated strong performance but may struggle with generalization to unseen data. 
In contrast, training-free methods leverage powerful pre-trained models like Large Language Models (LLMs) to address various downstream tasks with only a few examples. 
Nonetheless, training-free methods for DIE encounter two primary challenges: (1) understanding the complex relationship between layout and textual elements in VRDs, and (2) providing accurate guidance to pre-trained models. 
To address these challenges, we propose SAmple-centric In-context Learning (SAIL).
SAIL introduces a fine-grained entity-level textual similarity to facilitate in-depth text analysis by LLMs and incorporates layout similarity to enhance the analysis of layouts in VRDs.
Moreover, SAIL formulates a unified In-Context Learning (ICL) prompt template for various sample-centric examples, enabling tailored prompts that deliver precise guidance to pre-trained models for each sample.
Extensive experiments on FUNSD, CORD, and SROIE benchmarks with various base models (\eg, LLMs) indicate that our \method outperforms training-free baselines, even closer to the full-training methods, showing the superiority and generalization of our method. 
\end{abstract}

\begin{links}
    \link{Code}{https://github.com/sky-goldfish/SAIL}
\end{links}

\section{Introduction}\label{sec:intro}

\begin{figure}[t]
\centering
\includegraphics[width=1.0\columnwidth]{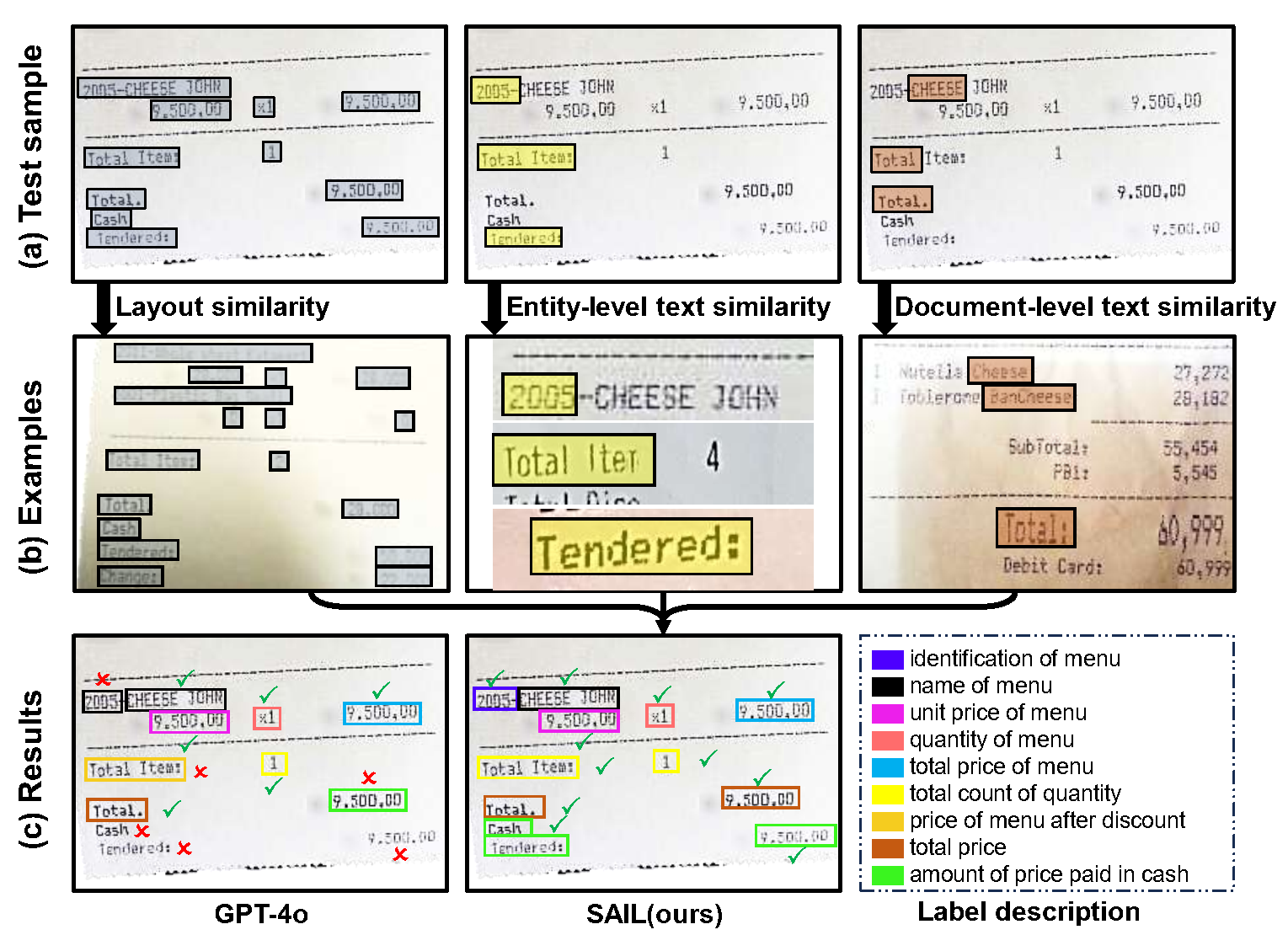}
\caption{
For the (a) test sample from the CORD dataset \cite{CORD}, our \method selects (b) layout similarity examples (grey marked), entity-level similarity examples (yellow marked), and document-level similarity examples (orange marked) to construct ICL prompts. 
(c) Benefiting from these examples, \method precisely extracts all information, while even the powerful GPT-4o \cite{gpt4o} misidentifies three entities and incorrectly labels three entities.
}
\label{teaser}
\end{figure}

Document Information Extraction (DIE) focuses on extracting structured information from Visually Rich Documents (VRDs) such as receipts, forms, and invoices~\cite{CORD, SROIE, FUNSD}.
Previous works, including LayoutLMv3~\cite{layoutlmv3}, primarily concentrate on full-training methodologies that demand extensive task-specific labeled data.
While these models have achieved notable success on the trained dataset, they often struggle to generalize effectively to unseen data, especially when the test data distribution significantly diverges from that of the training data. 
To address this challenge, training-free DIE methods~\cite{icld3ie} leverage powerful pre-trained models like Large Language Models (LLMs) that can generalize to unseen data given only a few examples, and thus begin to attract more research interests.

One of the primary challenges in the training-free DIE task is understanding the complex relationship between the document layout and its textual entities using only a few examples. 
VRDs possess discrete textual elements alongside flexible, inherently structured layouts, complicating the establishment of relationships between textual entities and the extraction of implicit layout information.
Even the advanced multi-modal LLMs like GPT-4o ~\cite{gpt4o} demonstrate limited effectiveness in performing DIE task. 
As illustrated in Figure \ref{teaser}(c), GPT-4o misidentifies three entity texts and labels three entity texts incorrectly, highlighting the challenges inherent in the training-free DIE task.

Another significant challenge is providing a clear and effective guidance to pre-trained models (\eg, LLMs). 
Although these models possess extensive knowledge and capabilities, they necessitate appropriate instructions for optimal performance on specific downstream tasks. 
Recent research has incorporated In-Context Learning (ICL) within LLMs to enhance the DIE performance~\cite{icld3ie}. 
This approach involves selecting a few textually similar examples and carefully crafting the in-context prompts with diverse demonstrations for the entire dataset. 
While this method shows promising results in GPT-3.5~\cite{ICL}, the fixed in-context examples fail to effectively guide different LLMs, leading to a significant performance decline when transitioning across different LLMs, as detailed in Table \ref{tab:compare_baseline}. 
To address these challenges, we propose a SAmple-centric In-context Learning (\method) method. 
Our method follows two core principles:  (a) To enhance LLMs' understanding of the complex interplay between layout and text within VRDs, the provided prompts must analyze the question from different angles in depth. 
(b) To ensure precise guidance, it is essential to develop a customized prompt for each test sample. 
Regarding the first principle, previous methods~\cite{icld3ie} only adopted rough document-level textual similarity for example selection, which inadequately supports LLMs in understanding textual information in lengthy documents.
Consequently, we propose a refined entity-level text similarity for in-depth text analysis.
Additionally, we incorporate layout similarity to identify examples that enjoy similar layouts, facilitating LLMs in comprehending complex layout information in VRDs.
The three distinct examples are illustrated in figure \ref{teaser}(b).
For the second principle, we select distinct examples for each test sample and integrate them into a unified prompt template with clear instructions to devise a tailored sample-centric in-context prompt.

Equipped with these designs, our proposed \method demonstrates versatility across various LLMs on multiple benchmarks. 
\method not only stably surpasses all training-free baselines, but even achieves comparable performance to many fully-trained models when implemented with GPT-4. 
Overall, our main contributions can be summarized as follows: 

\begin{itemize}
    \item 
    We introduce \textbf{layout similarity} and \textbf{entity-level text similarity}, each highlighting unique facets of VRDs,  resulting in a thorough and in-depth analysis of VRDs. 
    \item
    To form \textbf{sample-centric} in-context prompts, we propose
    a unified ICL prompt template applicable to various examples.  
    With clear instructions, LLMs enhance their attention to specific information in the examples.
    \item 
    We conduct extensive experiments on multiple benchmarks including FUNSD, CORD, and SROIE with various base LLMs. Our \method achieves superior performance than training-free baselines, even closer to the performance of full-training methods. 
\end{itemize}

\section{Related Works}\label{sec:works}

\textbf{Document Information Extraction} (DIE). 
Traditional DIE methods primarily rely on extensive datasets for model pre-training and subsequent fine-tuning on downstream tasks. 
These methods can be classified into four main categories.
The first category consists of grid-based methods~\cite{chargrid, cutie,bertgrid,visualwordgrid}, which encode each document page as a two-dimensional 
character grid of characters to preserve the document's layout.
The second category, graph-based methods, utilizes Graph Convolutional Networks (GCN) \cite{graphie, graph} or Graph Neural Networks (GNN) \cite{matchvie} for DIE. 
The third category encompasses transformer-based~\cite{transformer} methods.
Traditional methods design small models in specialized fields.
Some methods integrate text semantics and layout modality for model pre-training~\cite{structurallm, bros, lilt,wang2023adaptive}, while other methods jointly leverage text, layout, and image modality to enhance document understanding~\cite{visiongrid,layoutlm,layoutlmv2,layoutlmv3}. 
A recent trend has seen numerous studies employing LLMs' advanced language capabilities. ~\cite{wang2023docllm, perot2023lmdx, lu2024bounding, li2024enhancing, luo2024layoutllm, fujitake2024layoutllm}.
In contrast to the categories above that necessitate OCR for text and box recognition, the final category aims to bypass the OCR process and establish end-to-end models \cite{VIES, Donut,liu2024textmonkey, mao2024visually,hu2024mplug,abramovich2024visfocus}. 
Despite the notable performance of many methods, they demand retraining for specific downstream tasks.

\textbf{In-Context Learning} (ICL). 
\citet{ICL} discovered that pre-trained LLMs can address unseen tasks using only a few examples without weight updates through ICL. 
From then on, ICL has been widely adopted in question answering~\cite{vqa,mmhqa, sciQA}, multi-modal named entity recognition~\cite{cai2023context}, and dialogue improvement~\cite{dialoguesafety, dialogue}.

\textbf{ICL-based DIE}. 
ICL presents a viable approach for performing the DIE task with minimal examples. 
ICL-D3IE~\cite{icld3ie}, the first work to construct ICL prompts for DIE, utilizes diverse demonstrations through examples selected via text semantic search. 
Nonetheless, ICL-D3IE exhibits limited generalization capabilities to novel LLMs, primarily due to its reliance on fixed examples and handcrafted prompts.
Our method clearly distinguishes itself from this work. 
First, we dynamically select unique examples for each test sample, in contrast to ICL-D3IE's fixed examples. 
Second, we employ a unified template to construct prompts that can be generalized to various LLMs, while ICL-D3IE adopts specifically designed prompts that are less adaptable to new models.
Third, we demonstrate that relying solely on document-level text similarity is inadequate for identifying optimal examples, and thus introduce layout similarity and entity-level text similarity for enhanced performance. 
With these designs, our method achieves better results than ICL-D3IE across various base LLMs. 
\section{Methods}\label{sec:method}

\begin{figure*}[t]
\centering
\includegraphics[width=\textwidth]{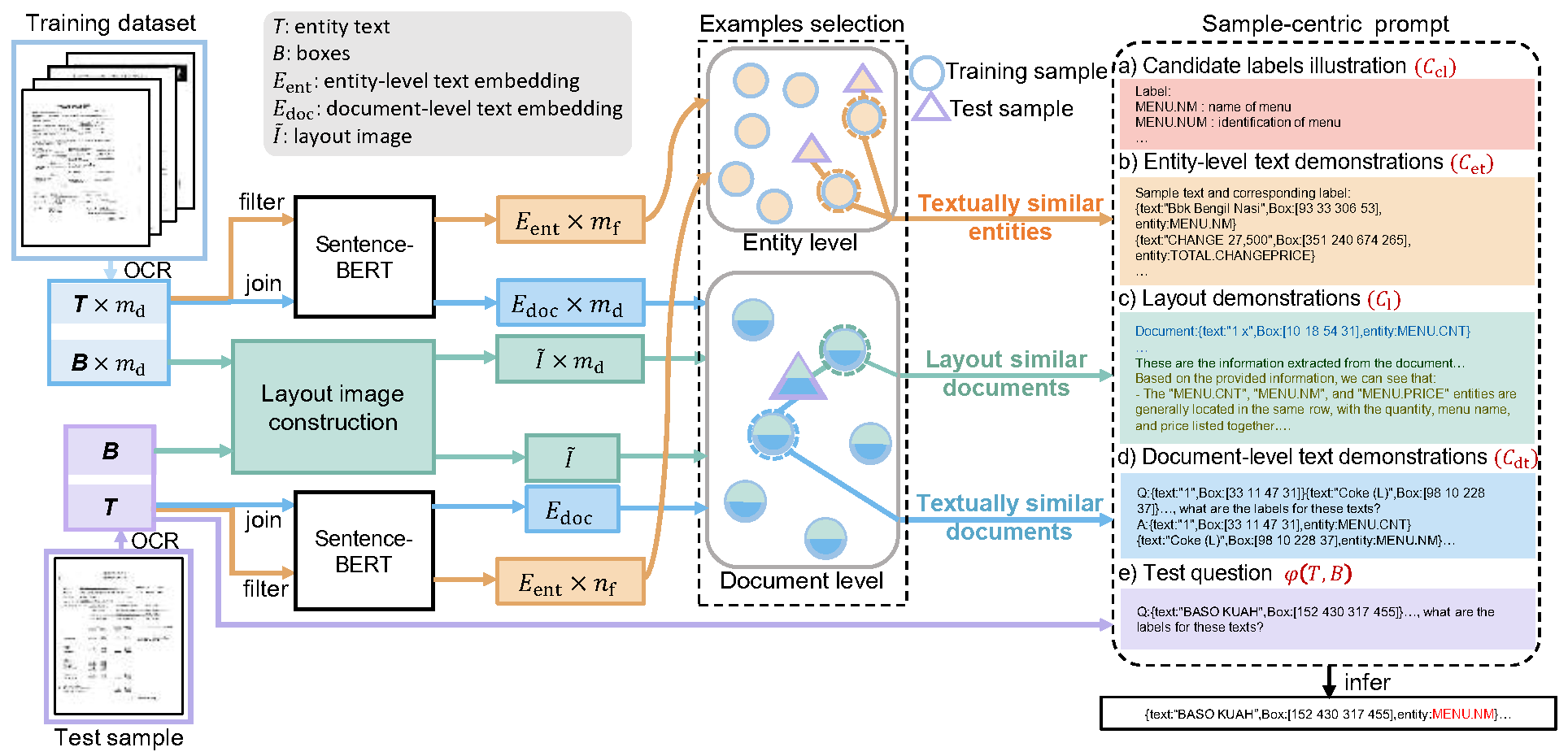} 
\caption{\textbf{Illustration of \method framework},
including extracting texts $T$ and boxes $B$ from document images, encoding them separately, selecting textually similar entities, layout similar documents, and textually similar documents for each test sample,  constructing sample-centric prompts using diverse examples, and generating predicted labels.
} 
\label{method fig}
\end{figure*}

\subsection{Problem Formulation}\label{subsec:preliminaries}
Training-free DIE leverages pre-trained models (\eg, LLMs) to extract specified categories of text information (\eg, company, address, and date~\cite{SROIE}) from VRDs. 
Specifically,  given a document image $I$,  the goal is to label all entities within $I$.  
First, entity texts $T=\{t_1,t_2,...,t_{n_{\rm e}}\}$ and their corresponding boxes $B=\{b_1,b_2,...,b_{n_{\rm e}}\}$ are recognized from $I$ by an OCR system, where $n_{\rm e}$ is the total number of entities in the document image. 
To effectively utilize LLMs, in-context prompts $C$ are designed to convey the extraction intention.
For ICL-based DIE, $C$ is constructed by selecting several examples demonstrating how to solve DIE tasks.
With these in-context prompts as illustrations, LLMs are tasked with generating labels $Y_{\rm pred}$ for all detected entities.
The process is achieved by maximizing the conditional probability $P(Y|T, B)$ while incorporating the prompts $C$ as an additional condition:
\begin{equation}
\label{eq1}
	P(Y|T,B)=\frac{1}{n_{\rm e}}\sum_{k=1}^{n_{\rm e}} P_{\rm LM}(l_k |C, \varphi(T,B)),
\end{equation}
where $P_{LM}$ is the conditional probability predicted by the LLMs, and $\varphi$ denotes the operation of converting the entity texts and boxes into a textual format suitable for LLMs' input. 
In training-free DIE, the construction of effective in-context prompt $C$ is crucial, which is the primary focus of this work.
Finally, the predicted labels $Y_{\rm pred}$ are evaluated using F1 scores against the ground truth labels $Y_{\rm gt}$.

\subsection{Overview Framework} 
To maximize $P(Y|T,B)$ with the in-context prompt 
$C$, we propose \method, a sample-centric in-context prompt construction method for DIE. 
\method focuses on designing $C$ for individual samples by automatically selecting tailored layout examples, document-level text similarity examples, and entity-level text similarity examples based on the test sample, subsequently leveraging these examples to generate $C$.

The overall architecture, illustrated in Figure \ref{method fig}, comprises five steps. 
Firstly, the test document image and $m$ training document images are processed through OCR to extract entity texts $T$ and boxes $B$.
Secondly, $T$ are transformed into entity-level text embeddings $E_{\rm ent}$ and document-level text embeddings $E_{\rm doc}$.
$B$ are used to construct layout image $\tilde{I}$.
Thirdly, $E_{\rm ent}$, $\tilde{I}$ and $E_{\rm doc}$ are used to select textually similar entities, layout similar documents, and textually similar documents for the test sample.
Then, these selections are substituted into the prompt template to form a tailored in-context prompt $C$.
Finally, LLM  performs inference with  $C$ and question $\varphi(T, B)$ to generate predicted labels $Y_{\rm pred}$.

\subsection{Document-Level Text Similarity Examples}
To improve the capability of ICL, we employ text semantic search to select the nearest training document examples for a given test sample~\cite{liu2022makes}.
The entity texts $T$ extracted from a document image are concatenated into a single sentence and encoded with Sentence-BERT~\cite{sentence-bert}, resulting in a text semantic embedding $E_{\rm doc}$ for the document.
We determine the nearest training examples by computing the document-level text similarity $T_{\rm sim \_ doc}$ between the test embedding $E_{\rm doc}^{\rm test}$ and $m$ training embeddings $E_{\rm doc}^{\rm train}$ using the cosine similarity score:
\begin{equation}
\label{cos}
	T_{\rm sim\_doc} = \frac{E_{\rm doc}^{\rm test} \cdot E_{\rm doc}^{\rm train}}{|| E_{\rm doc}^{\rm test}|| \; || E_{\rm doc}^{\rm train} ||}.
\end{equation}

\subsection{Entity-Level Text Similarity Examples}
The document-level text similarity $T_{\rm sim \_ doc}$ between a lengthy text document and the found text-similar documents is notably low. 
To facilitate LLMs in generating text with more relevant examples for learning, we propose entity-level text similarity examples, as shown in Figure \ref{method fig}.

Entity texts $T=\{t_1,t_2,\ldots,t_{n_{\rm e}}\}$ recognized by OCR are filtered to exclude texts consisting solely of numbers, which provide minimal semantic content.
Subsequently, the filtered $m_{\rm f}$ training entity texts and $n_{\rm f}$ test entity texts are encoded using Sentence-BERT to derive the semantic embedding $E_{\rm ent}$.
The entity-level text similarity $T_{\rm sim \_ ent}$ is computed from the semantic embedding $E_{\rm ent}$ by employing the cosine similarity score, defined as follows:
\begin{equation}
\label{entitysim}
	T_{\rm sim \_ ent} = \frac{E_{\rm ent}^{\rm test} \cdot E_{\rm ent}^{\rm train}}{|| E_{\rm ent}^{\rm test}|| \; || E_{\rm ent}^{\rm train} ||}.
\end{equation}

We select $n_{\rm s}$ textually similar entities for each test entity by nearest neighbor search and obtain $n_{\rm f}\times n_{\rm s} $ examples.

\subsection{Layout Similarity Examples}
\begin{figure}[t]
\centering
\includegraphics[width=\columnwidth]{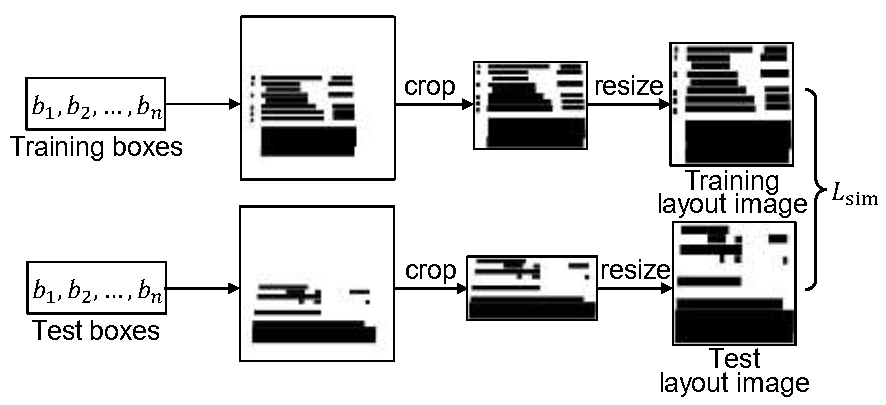} 
\caption{\textbf{Illustration of layout similarity evaluation}, including drawing boxes onto a blank image, cropping and resizing to form layout image, and comparing layout images.}
\label{layoutimage}
\end{figure}

To identify documents with similar layouts, we introduce a layout similarity assessment methodology, illustrated in  Figure \ref{layoutimage}.
Firstly, all $b_i$ from boxes $B=\{b_1,b_2,...,b_{n_{\rm e}}\}$ are rendered as black rectangles on a blank image.
Subsequently, we define the information area as the minimal region that contains all entity texts and crop the layout image to maintain a 10-pixel margin between the information area and the image borders.
Next, we standardize the layout image dimensions through resizing. 
Finally, we select $n_{\rm s}$ layout similar documents by calculating the layout similarity $L_{\rm sim}$ between the training layout image $\tilde{I}^{\rm train}$ and the test layout image $\tilde{I}^{\rm test}$ using Mean Square Error (MSE) loss:
\begin{equation}
\label{mse}
	L_{\rm sim} = \frac{1}{\rm MSE} = \frac{n_{\rm l}}{ (U - V)^{\rm T} (U - V)},
\end{equation}
where $U$, $V$ are the pixel matrix of $\tilde{I}^{\rm train}$ and $\tilde{I}^{\rm test}$, and $n_{\rm l}$ is the total number of pixels in the layout image.

Moreover, to enhance the understanding of layouts by LLMs, we substitute the boxes from the cropped image $B'$ for all documents in the prompt instead of using $B$.

\subsection{ Sample-Centric ICL Prompt Template}
To construct $C$ for an individual test sample,  we propose an adaptive sample-specific ICL prompt template.
The template is comprised of 5 parts: candidate labels illustration $C_{\rm cl}$, entity-level text demonstrations $C_{\rm et}$, layout demonstrations $C_{\rm l}$, document-level text demonstrations $C_{\rm dt}$ and test question $\varphi(T,B)$, as shown on the right of Figure~\ref{method fig}.

Candidate labels illustration $C_{\rm cl}$ enumerates all potential labels for the DIE task.
For abbreviated labels, a corresponding natural language description is appended.     

Entity-level text demonstrations $C_{\rm et}$ present textually similar entities. 
The prompt $p_{\rm e}$ ``\textit{Sample text and corresponding label:}" in conjunction with the labels of the selected $n_{\rm s}$ textually similar entity examples $Y_{\rm et}$, formulates the entity-level text similarity demonstrations:
\begin{equation}
\label{entity prompt}
	C_{\rm et}={\rm CONCAT}[p_{\rm e}, Y_{\rm et}].
\end{equation}

Layout demonstrations $C_{\rm l}$ aim to facilitate LLMs in analyzing the layout of the test document.
After obtaining $n_{\rm s}$ layout similar documents, we introduce a layout analysis step. 
This step enables LLMs to comprehend the overall document structure
 and the relationship between layout and label selection.
 The layout analysis prompt $p_{\rm a}$ is defined as:   ``\textit{These are the information extracted from the document through OCR, and the Box is the position of the text in the document. Please analyze where each label is generally located in the document.}", which can apply to any dataset.
 The labels of layout-similar documents $Y_{\rm l}$ are input into LLMs together with $p_{\rm a}$, allowing LLMs to analyze the layout information in layout-similar documents by themselves. 
The resulting output from the LLM is denoted as $A_{\rm l}$.
A layout similarity demonstration $C_{\rm l}$ is formulated as follows:
\begin{equation}
\label{layout prompt}
	C_{\rm l}={\rm CONCAT}[Y_{\rm l}, p_{\rm a}, A_{\rm l}].
\end{equation}

Document-level text demonstrations $C_{\rm dt}$ showcase textually similar documents in question-answer format, guiding LLMs to produce answers in a specific format. 
The textually similar documents $X_{\rm dt}$, the ground truth answer $Y_{\rm dt}$ and the DIE instruction $p_{\rm q}$ such as ``\textit{What are the labels for these texts?}" form the Document-level text demonstration prompt:
\begin{equation}
\label{DT}
	C_{\rm dt} = {\rm CONCAT} [ X_{\rm dt} , p_{\rm q} , Y_{\rm dt} ].
\end{equation}

Finally, the test question $\varphi(T,B)$ for the test sample is:
\begin{equation}
\label{Q}
	\varphi(T,B) = {\rm CONCAT}[ T, B' , p_{\rm q} ].
\end{equation}

\subsection{Inference}
After selecting a diverse set of examples, ICL prompts facilitate LLMs in generating entity labels $Y_{\rm pred}$. %
This process is mathematically represented as follows: %
\begin{equation}
\label{cp}
    P(Y|T,B)=\frac{1}{n_{\rm e}}\sum_{k=1}^{n_{\rm e}} P_{\rm LM}(l_k |C_{\rm cl}, C_{\rm et}, C_{\rm l}, C_{\rm dt}, \varphi(T,B)).
\end{equation} %
Subsequently, entity labels $Y_{\rm pred}$ are extracted from the generated output. 
We assess the accuracy of $Y_{\rm pred}$ against the ground truth labels $Y_{\rm gt}$ utilizing the F1 score.

\section{Experiments}\label{sec:exp}

\subsection{Datasets, Metrics, and Details}
\textbf{FUNSD}~\cite{FUNSD} is a dataset for understanding the content of tables in scanned documents.
It contains 149 tables and 7,411 entities in the training set, and 50 tables and 2,332 entities in the test set.
In the DIE task, the candidate labels of the FUNSD dataset include ``Header", ``Question", ``Answer", and ``Other".

\textbf{SROIE}~\cite{SROIE} is another scanned receipt understanding dataset, containing 626 receipts in the training set and 347 in the test set. 
The DIE task needs to extract ``company", ``date", ``address", and ``total" information.

\textbf{CORD}~\cite{CORD} is a receipt understanding dataset that contains 800 training data, 100 test data, and 100 validation data. 
This dataset features 30 detailed and hierarchical labels, much more than the above two datasets.

\textbf{Metrics}. Following previous works~\cite{icld3ie}, we adopt entity-level \textbf{F1 score}, \textbf{precision} and \textbf{recall} as metrics.

\textbf{Details}. 
We evaluate our method using three LLMs: the open-source ChatGLM3 \cite{chatglm3} and the closed-source GPT-3.5 \cite{gpt3.5} and GPT-4 \cite{gpt4o}. 
Specifically, we use the \texttt{chatglm3-6b-32k} version for ChatGLM3, \texttt{gpt-3.5-turbo} API version for GPT-3.5, and \texttt{gpt-4o} API version for GPT-4.
For GPT-3.5 and GPT-4o, we set the temperature parameter to 0 to enhance the reproducibility. 
In the case of GPT-4o, we only provide text prompts as input, while also testing its multi-modal capabilities by providing document images and clear task instructions. 
In our experiments, for each test document, we select four textually similar documents and four layout-similar documents as examples due to the limitation of prompt token number. 
Furthermore, for each filtered test entity, we choose four textually similar entity examples.

\subsection{Results on DIE Benchmarks}

\textbf{Baselines}. 
We compare our \method against baseline models including BERT~\cite{bert}, LiLT~\cite{lilt}, BROS~\cite{bros}, XYLayoutLM~\cite{xylayoutlm}, LayoutLM~\cite{xylayoutlm}, LayoutLMv2~\cite{layoutlmv2}, and LayoutLMv3~\cite{layoutlmv3} in both full-training and few-shot settings. 
We borrow their metrics from~\cite{icld3ie}. 
Training-free methods including standard ICL and ICL-D3IE~\cite{icld3ie} are also compared. 
ICL-D3IE only reports the performance of standard ICL and ICL-D3IE with GPT-3.5, so we evaluate their performance with GPT-4 and ChatGLM3 using their official repositories.

\textbf{Quantitative results} are presented in Table \ref{tab:compare_baseline}. 
First, overall, our method stably outperforms ICL-D3IE across different LLMs on all datasets. 
Second, when switching the LLM from GPT-3.5 to ChatGLM3, the performance drop of ICL-D3IE is significantly larger than our \method (\eg, -73.8\% \vs  -12.73\% in SROIE), demonstrating that our method has better robustness and adaptability to various LLMs. 
Third, the performance of ICL-D3IE degrades slightly when transitioning from GPT-3.5 to the more advanced GPT-4 on FUNSD and SROIE datasets, further indicating its incompatibility with new LLMs. 
However, in all datasets, our method achieves better performance on more advanced GPT-4 than on GPT-3.5, which is intuitive and reasonable. 
These results demonstrate the advantages of our method.

\textbf{Qualitative results} are illustrated in Figure \ref{expbaseline}.
ICL-D3IE incorrectly predicts the entities on the three left green boxes as ``answer", while our \method accurately identifies them as ``question". 
This indicates that fixed examples in ICL-D3IE are insufficient to guide LLMs in effectively learning the relationship between discrete texts, highlighting the importance of selecting diverse examples for each test sample.

\begin{table}[t]
\small
\setlength\tabcolsep{1.5pt}
\centering
\begin{tabular}{c|c|c|cccc}
\toprule
    Setting & \multicolumn{2}{c|}{Methods} & FUNSD & CORD & SROIE \\
    \midrule
    \multirow{7}{*}{\makecell[c]{Full-Training}} & \multicolumn{2}{c|}{BERT$_{\rm BASE}$}& 60.26& 89.68 & 90.99\\
     & \multicolumn{2}{c|}{LiLT$_{\rm{BASE}}$} & 88.41&  96.07 & 94.68 \\
    & \multicolumn{2}{c|}{BROS$_{\rm{BASE}}$}  & 83.05& 95.73 & 95.48 \\
    & \multicolumn{2}{c|}{XYLayoutLM$_{\rm{BASE}}$} & 83.35& 94.45 & 95.74\\
    & \multicolumn{2}{c|}{LayoutLM$_{\rm{BASE}}$} & 79.27 & 91.06 & 94.38 \\
    & \multicolumn{2}{c|}{LayoutLMv2$_{\rm{BASE}}$} & 82.76& 94.95 &  96.25   \\
    & \multicolumn{2}{c|}{LayoutLMv3$_{\rm{BASE}}$} & 90.29 & 96.56 & 96.89 \\
    \midrule
    \multirow{7}{*}{Few-Shot} & \multicolumn{2}{c|}{BERT$_{\rm{BASE}}$}  & 38.76& 38.88 & 38.76\\
    & \multicolumn{2}{c|}{LiLT$_{\rm{BASE}}$}  & 54.88&  69.12 & 84.03 \\
    & \multicolumn{2}{c|}{BROS$_{\rm{BASE}}$}& 59.46& 72.78 & 76.78 \\
    &  \multicolumn{2}{c|}{XYLayoutLM$_{\rm{BASE}}$}  & 65.44& 69.16 & 75.66 \\
    &\multicolumn{2}{c|}{LayoutLM$_{\rm{BASE}}$} & 32.49 & 40.19 & 76.79 \\
    & \multicolumn{2}{c|}{LayoutLMv2$_{\rm{BASE}}$} & 71.42 & 65.71 &  81.81   \\
    & \multicolumn{2}{c|}{LayoutLMv3$_{\rm{BASE}}$} & 70.67 & 70.13 & 79.13 \\
     \midrule
     \multirow{9}{*}{Training-Free}&\multirow{3}{*}{ChatGLM3}  &Standard ICL & 40.93 & 67.30 & 81.37 \\
     & & ICL-D3IE & 35.90 & 36.44 & 18.83 \\
     & & \method (ours) & \textbf{58.24} & \textbf{83.04} & \textbf{85.03} \\
      \cmidrule{2-6}
     &\multirow{3}{*}{GPT-3.5}  & Standard ICL & 72.76 & 68.34 & 82.11\\
     & & ICL-D3IE & \textbf{83.66} & 87.13 & 92.63 \\
     & & \method (ours) & 83.48 & \textbf{95.80} & \textbf{97.76} \\
     \cmidrule{2-6}
     &\multirow{3}{*}{GPT-4}  & Standard ICL &  75.15& 90.22 &  96.00 \\
     & & ICL-D3IE & 78.94 & 87.47 & 89.23\\
     & & \method (ours) &  \textbf{84.67}& \textbf{96.41}  & \textbf{98.18} \\
\bottomrule
\end{tabular}
\caption{
    \textbf{Quantitative results} with F1 metric. 
    Our \method stably surpasses baselines across various base LLMs. 
}
\label{tab:compare_baseline}
\end{table}

\subsection{Comparison with Multi-modal LLMs}

\textbf{Baselines}. 
Recent years have witnessed the rapid development of multi-modal LLMs (MLLMs) represented by GPT-4o~\cite{gpt4o}. 
To further validate the effectiveness of our method, we also compare our \method with MLLMs including open-source LLaVA-1.5~\cite{llava1.5} and proprietary GPT-4o. 
We provide these MLLMs with explicit and detailed instructions to inform the task definition.

\textbf{Quantitative results} are provided in Table \ref{tab:compare_mllm}. 
The open-source LLaVA exhibits limited DIE capabilities, resulting in a low F1 score (e.g., 0.7\% in FUNSD). 
The proprietary GPT-4o significantly outperforms LLaVA (50.72\% vs 0.7\% in FUNSD), yet still falls short when compared to specialized DIE methods.
Therefore, despite their rapid evolution, MLLMs still underperform in the DIE task, highlighting the importance and contribution of our proposed work.

\begin{table*}[!ht]
\begin{minipage}[t]{1.0\textwidth}
\centering
\setlength\tabcolsep{7.5pt}
\begin{tabular}{c|ccc|ccc|ccc}
\toprule
     \multirow{2}{*}{Methods}& \multicolumn{3}{|c|}{SROIE} & \multicolumn{3}{|c|}{CORD}  & \multicolumn{3}{|c}{FUNSD}  \\
    \cmidrule{2-10}
    & F1 & Precision & Recall & F1 & Precision & Recall & F1 & Precision & Recall\\
    \midrule
    GPT-4o & 47.49 & 46.77 & 48.24 & 
    71.53 & 82.96 & 62.87 & 
    50.72 & 73.01 & 38.85 \\
    LLaVA-v1.5-7B & 2.32 & 5.49 & 1.47 & 
    8.85 & 67.39 & 4.74 & 
    0.70 & 61.54 & 0.35 \\
    \midrule
    \method (ours) & \textbf{98.18} & \textbf{97.72} & \textbf{98.64}   &\textbf{96.41} & \textbf{96.41} &\textbf{96.41}   & \textbf{84.67} & \textbf{84.67} & \textbf{84.67} \\
\bottomrule
\end{tabular}
\caption{
\textbf{Performance comparison} with multi-modal LLMs. 
Multi-modal LLMs even powerful GPT-4o still struggle with DIE tasks and our method significantly surpasses GPT-4o and LLaVA-v1.5-7B. 
}
\label{tab:compare_mllm}
\vspace{10pt}
\end{minipage}
\vfill
%
\begin{minipage}[t]{0.48\textwidth}
\small
\setlength\tabcolsep{8pt}
\centering
\begin{tabular}{c|ccc}
\toprule
   Setting & FUNSD & CORD & SROIE \\
    \midrule
    Fixed example &                  74.23  & 82.35&  91.08 \\
    Adaptive example &          \textbf{83.48} & \textbf{95.80} & \textbf{97.76} \\
    \midrule
    $\Delta$ & 9.25 & 13.45 & 6.68\\
\bottomrule
\end{tabular}
\caption{
    \textbf{Ablation study of adaptive examples} with F1 metric. 
    Adaptive examples is superior than fixed examples. 
}
\label{tab: fix}
\end{minipage}
\hfill
\begin{minipage}[t]{0.48\textwidth}
\small
\renewcommand{\arraystretch}{1.1}
\setlength\tabcolsep{12pt}
\centering
\begin{tabular}{c|cc}
\toprule
    \diagbox [width=8em,trim=l] {Text}{Layout} & Ascending & Descending \\
    \midrule
    Ascending & 95.19 & 94.73 \\
    Descending & 94.73 & \textbf{95.80} \\
\bottomrule
\end{tabular}
\caption{
    \textbf{Performance comparison of the example order} in the prompt with F1 metric in the CORD dataset. 
}
\label{tab: order}
\end{minipage}
\vspace{10pt}
\vfill
\begin{minipage}[t]{0.48\textwidth}
\centering
\includegraphics[width=1.0\columnwidth]{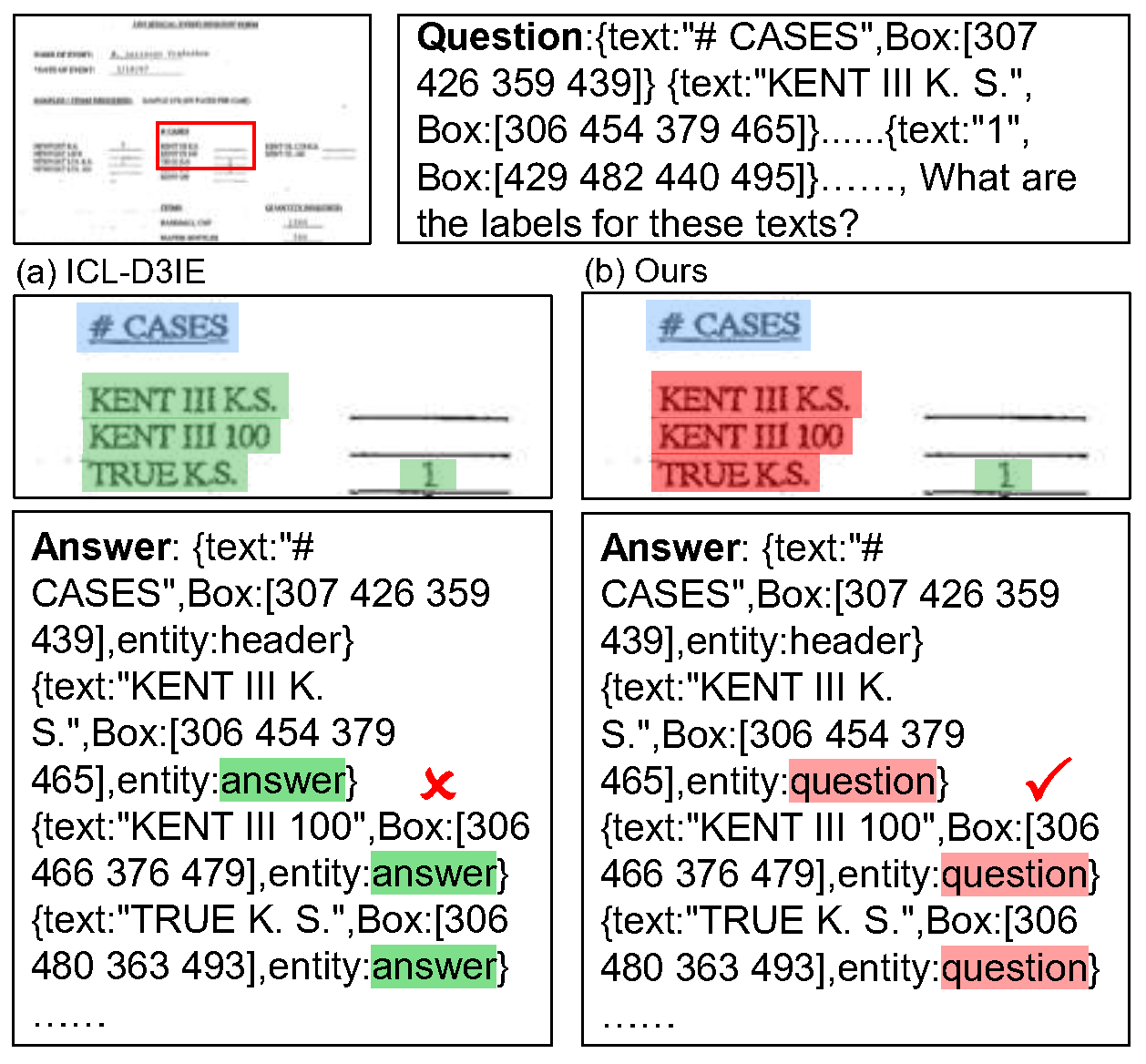}
\captionof{figure}{\textbf{Case study} on performance comparison of (a) ICL-D3IE and (b) our \method. ICL-D3IE wrongly predicts the three green boxes on the left as ``answer". In contrast, our proposed \method correctly predicts them as ``question".}
\label{expbaseline}
\end{minipage}
\hfill
\begin{minipage}[t]{0.48\textwidth}
\vspace{-222pt}
\begin{minipage}[t]{1.0\textwidth}
\small
\setlength\tabcolsep{2.5pt}
\centering
\begin{tabular}{c|ccc|ccc}
\toprule
    \multirow{2}{*}{\#} & \multicolumn{3}{c|}{Similar} & \multirow{2}{*}{FUNSD} & \multirow{2}{*}{CORD} & \multirow{2}{*}{SROIE} \\
     & Text-Doc. & Layout & Text-Ent. & & & \\
    \midrule
    0 &  &  & &                      69.87  & 83.04&  95.08 \\
    1 & $\checkmark$ &  & &           69.60 & 92.13&  96.38 \\
    2 & $\checkmark$ & $\checkmark$ & &   73.13   & 92.97&  97.24 \\
    3 & $\checkmark$ &  & $\checkmark$&     81.67  & 92.51&  97.13 \\
    4 & $\checkmark$ &$\checkmark$  & $\checkmark$& \textbf{83.48} & \textbf{95.80} & \textbf{97.76}   \\
\bottomrule
\end{tabular}
\caption{
    \textbf{Ablation study of various similarity} with F1 metric. 
    Text-Doc., Layout, \& Text-Ent. mean textual similar documents, layout similar documents, \& textual similar entities. 
}
\label{tab: main ablation}
\end{minipage}
\vfill
\begin{minipage}[t]{1.0\textwidth}
\centering
\includegraphics[width=0.81\columnwidth]{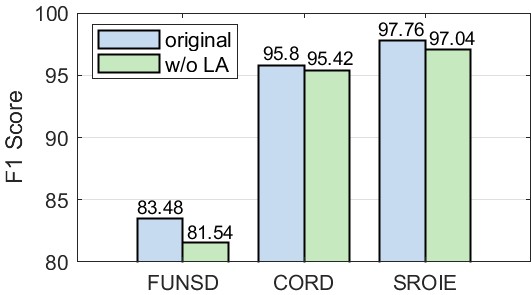}
\captionof{figure}{\textbf{Ablation study of layout analysis}. ``w/o LA" means without adding layout analysis. Adding layout analysis achieves higher F1 scores across all three datasets.}
\label{layoutanylize}
\end{minipage}
\end{minipage}
\end{table*}

\subsection{Ablation Studies}

\textbf{Effect of Adaptive Example}.
We assess the influence of adaptive examples by employing both fixed and adaptive examples to construct in-context prompts within the same prompt template. 
The base LLM is selected as GPT-3.5, and the results are illustrated in Table \ref{tab: fix}. 
The utilization of adaptive examples results in superior F1 scores, confirming the effectiveness of our method. 
Among the three datasets, the performance improvement with adaptive examples is most pronounced in the CORD dataset (13.45\%). 
Note that the CORD dataset contains 30 labels, much more complex than the other two datasets with only four labels. 
This suggests that sample-centric examples could more effectively guide the LLMs to comprehend the layout and text information especially in complex situations.

\begin{figure*}[t]
\centering
\includegraphics[width=0.95\textwidth]{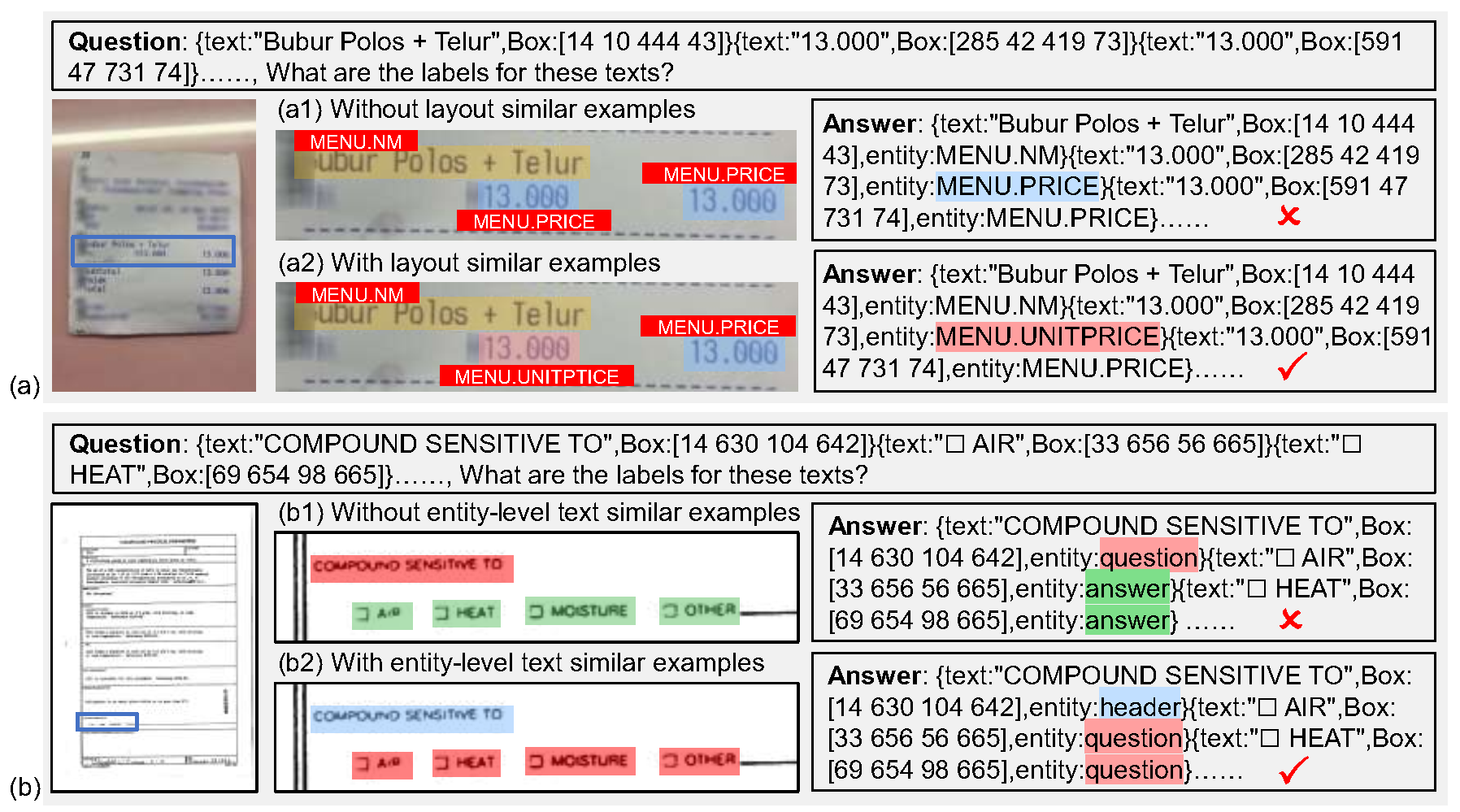}
\caption{
(a) \textbf{Case study} on comparison of (a1) without and (a2) with layout similar examples. Adding layout similar examples helps accurately distinguish between the two ``13.000".
(b) \textbf{Case study} on comparison of (b1) without and (b2) with entity-level text similar examples. The LLM labels entities correctly with the demonstration of entity-level text similar examples. 
} 
\label{casestudy}
\end{figure*}

\textbf{Effect of Different Examples}.
We conduct ablation experiments using GPT-3.5 to evaluate the influence of different examples, as shown in Table \ref{tab: main ablation}. 
In \#0, where none of the three examples are available, we employ fixed random examples to instruct the LLM to generate answers in a specific format, simplifying the label extraction.
The highest F1 score is observed when document-level text similarity examples, layout examples, and entity-level text similarity examples (\#4) are used, validating the efficiency of the three examples.
The addition of layout examples (\#1 \vs \#2) or entity-level text similarity examples (\#1 \vs \#3) to document-level text similarity examples results in superior F1 scores.

For the long text FUNSD dataset, the F1 score with document-level text similarity examples is even lower than fixed examples (\#0 \vs \#1). 
This could be attributed to the inherent randomness of LLM generation, but it also signifies that in lengthy documents, document-level text similarity examples do not provide effective guidance for the LLM.
In the FUNSD dataset, adding entity-level text similarity examples (10.35\%, \#2 \vs \#4) is much superior than adding layout similarity examples (1.81\%, \#3 \vs \#4), suggesting that entity-level text similarity examples are more important for lengthy documents.
For the CORD and SROIE datasets, removing layout similarity examples (\#3 \vs \#4) causes a greater F1 score decrease than omitting entity-level text similarity examples (\#2 \vs \#4), indicating the higher significance of layout information for these two datasets.

\textbf{Effect of Example Order}.
We perform experiments on the CORD dataset using GPT-3.5 to test the effect of example order, as detailed in Table \ref{tab: order}.
When layout-similar and text-similar document examples are arranged in a consistent order based on their similarity, F1 scores tend to be higher.
This phenomenon may result from improved attention allocation within the LLM due to the consistent ordering.
Furthermore, the highest F1 scores are observed when layout-similar and text-similar examples are sorted from high to low similarity concerning the test sample.
This suggests that the LLM can capitalize on the information presented first.

\textbf{Effect of Layout Analysis}. 
Our methodology requires the LLM to perform layout analysis on our searched layout-similar examples. 
To assess the impact of layout analysis, we conduct comparative experiments with / without the layout analysis on the FUNSD, CORD, and SROIE datasets using the GPT-3.5. 
As illustrated in Figure \ref{layoutanylize}, the results indicate that F1 scores are consistently higher when incorporating layout analysis compared to only using layout-similar examples across all datasets.
This suggests that layout analysis is able to enhance the LLM's comprehension of layout.

\textbf{Case Study}.
Figure \ref{casestudy}(a) illustrates a comparison from the CORD dataset regarding the inclusion of layout demonstrations in the prompt. 
Using the prompt without layout similar demonstrations, the LLM predicts two ``13.000" both as ``MENU.PRICE”, while our \method distinguishes the left ``13.000" as ``MENU.UNITPRICE” and the right ``13.000" as ``MENU.PRICE”. 
This outcome underscores the necessity of incorporating layout demonstrations for LLMs to grasp document structure effectively. 
Figure \ref{casestudy}(b) showcases a comparison from the FUNSD dataset about the addition of entity-level text demonstrations in the prompt. 
Upon omitting these demonstrations, the LLM mistakenly predicts “COMPOUND SENSITIVE TO” as “question” and incorrectly classifies the four subsequent entities as “answer”. 
Although this prediction makes sense in terms of layout, it fails to correspond with the textual context, highlighting the critical role of entity-level text similarity examples.

\section{Conclusions and Limitations}\label{sec:conclusion}

In this work, we propose \method, a sample-centric ICL method for training-free DIE task.
Our \method leverages layout similarity and entity-level text similarity in combination with a unified prompt template, constructing tailored prompts for each test sample, showcasing superiority over baselines on three DIE benchmarks with different LLMs. 
\section{Acknowledgments}\label{sec:ack}
This work was supported by the National Natural Science Foundation of China (No. 62422311, 62176152), and Shanghai Committee of Science and Technology, China(No. 24TS1413500).

{\small
\bibliography{aaai25}
}


\appendix
\renewcommand\thefigure{A\arabic{figure}}
\renewcommand\thetable{A\arabic{table}}  
\renewcommand\theequation{A\arabic{equation}}
\setcounter{equation}{0}
\setcounter{table}{0}
\setcounter{figure}{0}

\section*{Appendix}

\section{Implementation Detail} \label{supp:sec:implementation}

We test our \method with ChatGLM3 on an RTX 3090 GPU.
For FUNSD, we set the maximum token output to 2500;
For CORD and SROIE, we set the maximum token output to 1500.
Due to the randomness of the generation of LLMs, we have set some simple checks, such as whether there is a ``\{", and if the generated result does not meet the requirements, it is regenerated. 
In the FUNSD dataset, we set the number of examples to 2, and if the number of tokens exceeds the limit, the number of examples is changed to 1.
In the SROIE dataset, we set the number of examples to 4, and if the number of tokens exceeds the limit, the number of examples is changed to 2.
\section{Additional Ablation Studies} \label{supp:sec:ablation}

\textbf{Results on Synthetic Data}.
We explore two data-synthetic methods. 

\begin{itemize}
    \item 
    Replace text. Randomly replace the text with other texts of the same label while keeping the original layout.
    \item
    Replace layout. Keep the text unchanged and replace the layout with those from other documents. 
\end{itemize}

We conducted experiments on the CORD with gpt-3.5-turbo in Table \ref{tab: synthetic data}. Even with only synthetic data, SAIL achieves better results than GPT-4o and similar results with ICL-D3IE.

\textbf{Constructing an Example Pool}.
Because the extracted information of different datasets varies, we explored how to construct an example pool for each dataset. We found that even using only 30 examples, the performance is still high, as shown in Table \ref{tab: common pool}.

\textbf{Effect of the Layout Similarity Comparison Method}. 
We present the F1 score of different layout similarity comparison methods in Table \ref{tab: similarity method}.
When using cosine similarity, the layout images are pulled into a one-dimensional vector before comparing similarity.
From the table, it can be seen that the F1 score of using Mean Square Error (MSE) to compare the similarity between layout images is the highest.
Besides, the F1 score of comparing the cosine similarity is much lower than the F1 score of comparing the MSE or Structural Similarity (SSIM) between the images. 
This indicates that pulling into a one-dimensional vector destroys the overall structure of the layout image and loses some layout information. 
Therefore, it is necessary to directly compare the similarity between the layout images.

\begin{table}[t]
    \setlength\tabcolsep{8pt}
    \centering
    \begin{tabular}{c|c|c}
    \toprule
    Method & Data & F1 Scores \\
    \midrule
    GPT-4o &   -  &    50.72 \\
    ICL-D3IE &   100\% real  & 87.13   \\
    SAIL &   100\% real  & 95.80  \\
    SAIL &   50\% real  & 92.59   \\
    SAIL &   50\% replace text &  86.71  \\
    SAIL &   50\% replace layout  & 85.10  \\
    \bottomrule
    \end{tabular}
    \caption{\textbf{Results on synthetic data} with F1 metric in the CORD dataset.}
    \label{tab: synthetic data}
\end{table}

\begin{table}[t]
    \setlength\tabcolsep{8pt}
    \centering
    \begin{tabular}{c|c|c}
    \toprule
    CORD, F1 & GPT-3.5-turbo  & GPT-4o  \\
    \midrule
    full dataset &  95.80  & 96.41  \\
    half dataset &  92.59  & 94.65 \\
    30 examples & 84.57 & 92.36 \\
    \bottomrule
    \end{tabular}
    \caption{\textbf{Results on constructing an example pool} with F1 metric in the CORD dataset.}
    \label{tab: common pool}
\end{table}

\begin{table}[t]
    \setlength\tabcolsep{8pt}
    \centering
    \begin{tabular}{c|c}
    \toprule
    Layout Similarity Comparison Methods & F1 Scores \\
    \midrule
    Cosine Similarity &         94.73 \\
    Structural Similarity (SSIM) &      95.42    \\
    Mean Square Error (MSE) & \textbf{95.80} \\
    \bottomrule
    \end{tabular}
    \caption{\textbf{Performance comparison of the layout similarity comparison methods} with F1 metric in the CORD dataset.}
    \label{tab: similarity method}
\end{table}

\begin{figure}[t]
\centering
\includegraphics[width=1.0\columnwidth]{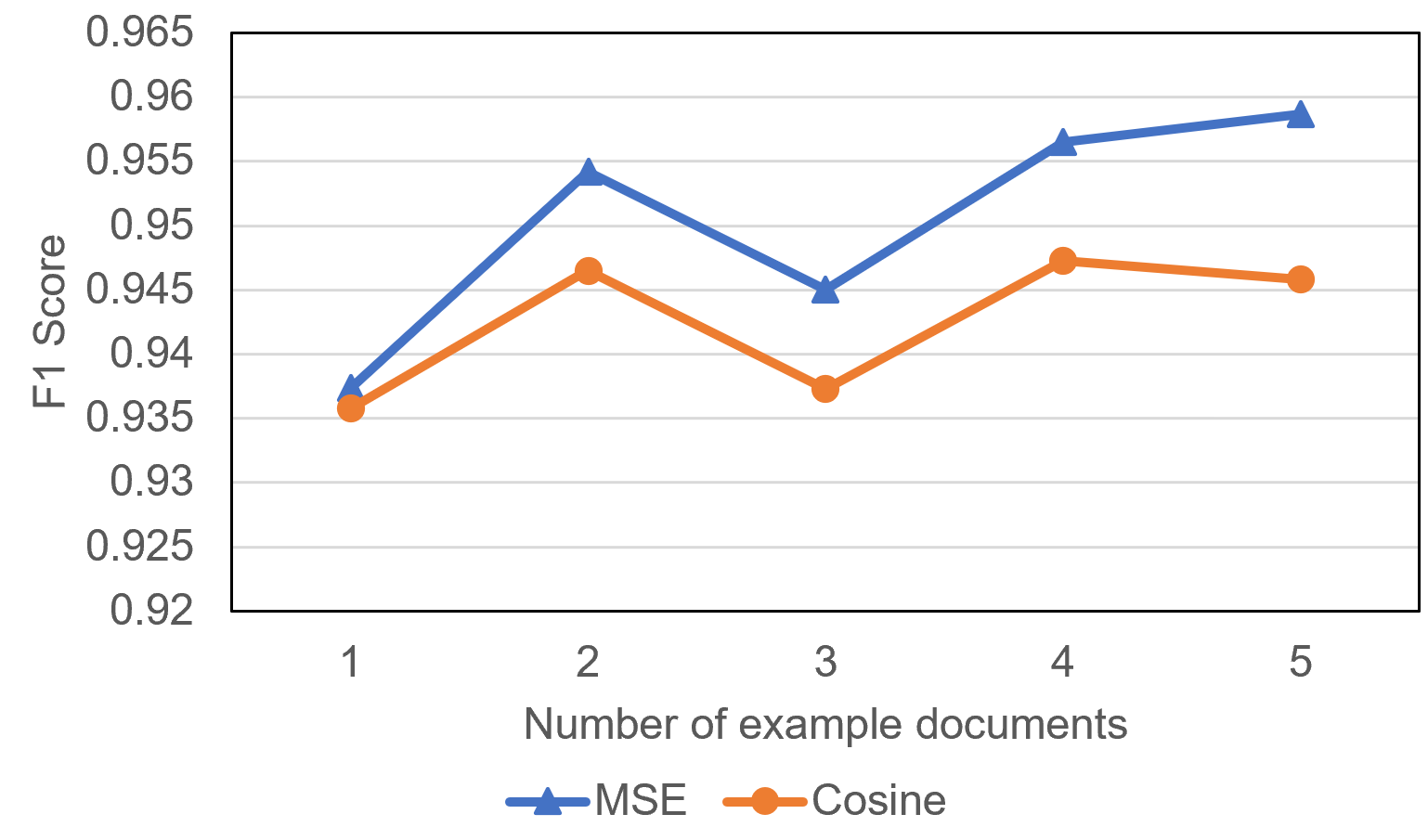}
\caption{\textbf{Performance comparison of the number of example documents} with F1 metric in the CORD dataset.
}
\label{number document}
\end{figure}

\textbf{Effect of the Number of Document-Level Examples}.
We tested the effect of the number of document-level examples in the CORD dataset when comparing the similarity of layout images using MSE and cosine similarity separately.
In our experience, the number of layout-similar document examples and textually similar document examples were the same and this number was the independent variable.
Due to the limitation on the length of the prompt words, only the cases with 1–5 document-level examples were tested. 
The results are shown in Figure \ref{number document}. 
It can be observed that, overall, the more document-level examples selected, the higher the F1 score is. 
However, when the number of document-level examples is 3, the prediction effect is significantly worse compared to when the number of document-level examples is 2 and 4. 
This is a very peculiar phenomenon. 
Additionally, the performance difference when the number of document-level examples is 2 and 4 is not very large. 
If cost-saving is a priority, using shorter prompt words and selecting 2 layout-similar document examples and 2 textually similar document examples can also yield good results.

\textbf{Effect of Different Resize Methods}.
When comparing the similarity of layout images processed by SSIM, the effect of different methods of resizing layout images was tested in the CORD dataset.
Our experience includes 5 resize methods.
The first method involves calculating the position coordinates of the boxes in the resized image before drawing them. 
Since the coordinates can only be integers, they need to be rounded off first.
The resized layout images are generated using these coordinates directly.
The second to fourth methods involve bilinear interpolation, LANCZOS interpolation, and area interpolation.
Using these methods, the layout image is no longer binary after resizing. 
The fifth method involves first performing LANCZOS interpolation, and then binarizing the image with 128 as the boundary to obtain a black-and-white binary layout image.
The result is shown in Table \ref{tab: resize}. 
From the table, it can be seen that the F1 scores of obtaining non-binary layout images through bilinear interpolation, area interpolation, and LANCZOS interpolation are somewhat lower. 
Among them, area interpolation still generates binary layout images by choosing neighboring pixel values when enlarging images, and generates non-binary layout images when reducing images, so its F1 score is relatively higher than bilinear interpolation and LANCZOS interpolation. 
Therefore, comparing the similarity of images by generating binary layout images after resizing is more conducive to correct prediction when searching for similar layout documents. 
Among the two ways to generate binary layout images, the F1 score of LANCZOS interpolation and binarization is a bit higher, indicating that it is the best way to resize layout images.

\begin{table}[t]
    \centering
    \begin{tabular}{c|c}
    \toprule
    Resize Methods & F1 Scores \\
    \midrule
    Calculate coordinates before generation & 95.26 \\
    \midrule
    Bilinear Interpolation &         94.73 \\
    LANCZOS Interpolation&      94.88    \\
    Area Interpolation &  95.11 \\
    \midrule
    LANCZOS Interpolation and Binarization   &   \textbf{95.42}  \\
    \bottomrule
    \end{tabular}
    \caption{\textbf{Performance comparison of resize methods} with F1 metric in the CORD dataset.}
    \label{tab: resize}
\end{table}

\textbf{Effect of Different Boxes in the Prompt}.
In \method, we substitute the boxes from the cropped image $B'$
for all documents in the prompt instead of using $B$.
To test the effect of this step, we conduct comparative experiments using two document-level examples and comparing the similarity of layout images with MSE.
The result is shown in Table \ref{tab: box}. 
The F1 score of using  $B'$ is higher,  proving that using the boxes from the cropped image is more beneficial for LLMs to observe the layout relationships between documents.

\begin{table}[t]
    \setlength\tabcolsep{18pt}
    \centering
    \begin{tabular}{c|c}
    \toprule
    Box From & F1 Scores \\
    \midrule
    Original image ($B$) & 95.11 \\
    Cropped image ($B'$) &        \textbf{95.42} \\
    \bottomrule
    \end{tabular}
    \caption{\textbf{Performance comparison of different boxes in the prompt} with F1 metric in the CORD dataset.}
    \label{tab: box}
\end{table}

\textbf{Effect of Representation Format for Textually  Similar Entities}.
Since the focus of the entity-level text similarity example is primarily on textual information, two representations are considered: one is "text: label", and the other is the format ``{text: ``...", Box: [$x_1$,$y_1$,$x_2$,$y_2$], entity: ...}".
Under the condition of selecting 2 entity-level text similarity examples for each testing entity, we conduct experiment to test the effect of two representation formats in the FUNSD dataset.
The result is shown in Table \ref{tab: format}.
It can be seen that the F1 score is higher when the position coordinates of the text blocks are included.
This indicates that in entity-level similar text demonstrations, the position coordinates of the text blocks are very important.
They can help LLMs make better judgments about which example to primarily refer to when the texts are the same but the labels are different.

\begin{table}[t]
    \centering
    \begin{tabular}{c|c}
    \toprule
    Representation Format & F1 Scores \\
    \midrule
    ``text: label" & 79.43 \\
   ``{text: ``...", Box: [$x_1$,$y_1$,$x_2$,$y_2$], entity: ...}" &         \textbf{82.20} \\
    \bottomrule
    \end{tabular}
    \caption{\textbf{Performance comparison of representation formats for textually similar entities} with F1 metric in the FUNSD dataset.}
    \label{tab: format}
\end{table}

\textbf{Effect of the Number of Entity-Level Text Similarity Examples}.
When using the format of ``{text:``…”, Box:[$x_1$,$y_1$,$x_2$,$y_2$], entity:…}”, we compared the effect of the number of entity-level text similarity examples in the FUNSD dataset. 
From Table \ref{tab: number entity}, it can be seen that increasing the number of examples can improve the F1 score.

\begin{table}[t]
    \setlength\tabcolsep{18pt}
    \centering
    \begin{tabular}{c|c}
    \toprule
    Number of Examples & F1 Scores \\
    \midrule
    2 & 82.20 \\
    4 &         \textbf{83.48} \\
    \bottomrule
    \end{tabular}
    \caption{\textbf{Performance comparison of the number of entity-level text similarity examples} with F1 metric in the FUNSD dataset.}
    \label{tab: number entity}
\end{table}

\section{Results}\label{supp:sec:result}

We use the F1 scores of the samples obtained by D3IE and our \method for Wilcoxon signed-rank, and the p scores are shown in Table \ref{tab: t-test}. The results show that the p-values of all three datasets are less than 0.05, indicating that the change is significant.

\begin{table}[ht]
    \centering
    \begin{tabular}{c|c c c}
    \toprule
    Datasets & FUNSD &CORD & SROIE \\
    \midrule
    p-value & 2.415e-2 &4.057e-5& 2.288e-17\\
    \bottomrule
    \end{tabular}
    \caption{\textbf{Wilcoxon signed-rank in D3IE and our \method}. The p-values of all three datasets are less than 0.05, indicating that the change is significant. }
    \label{tab: t-test}
\end{table}

\begin{figure*}[t]
\centering
\includegraphics[width=0.95\textwidth]{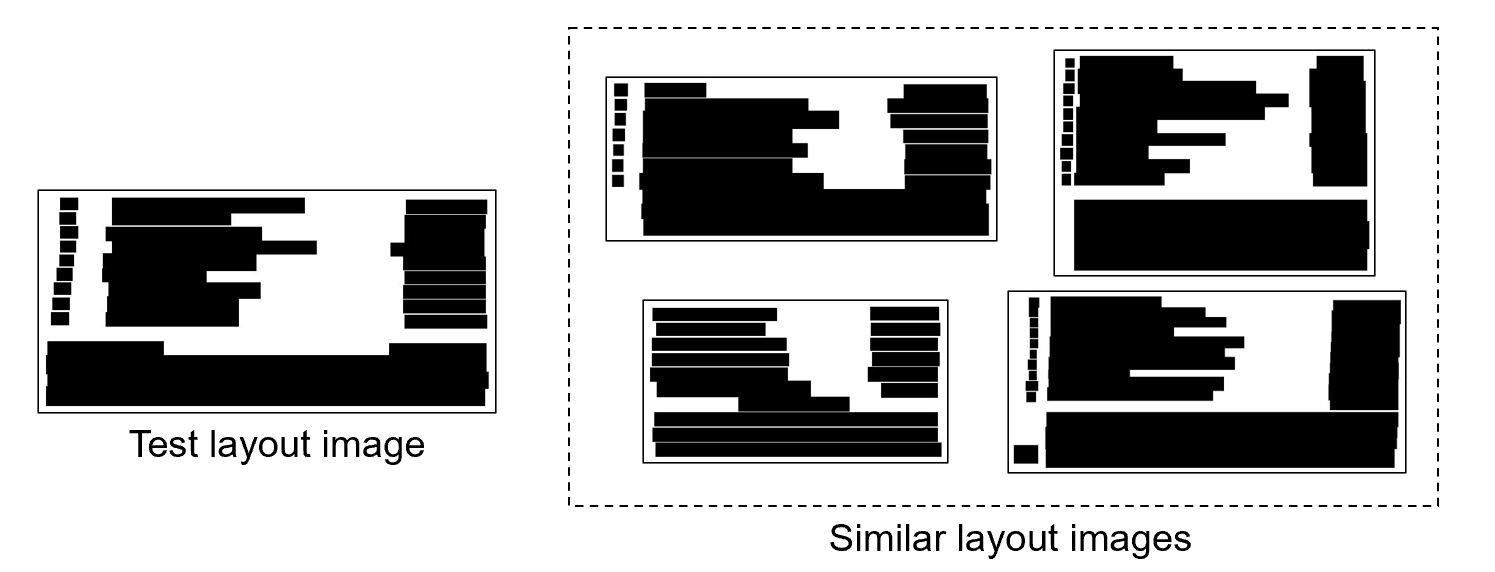}
\caption{
Example of the test layout image and selected similar layout images on CORD.
} 
\label{layoutimagecase}
\end{figure*}

\begin{figure*}[t]
\centering
\includegraphics[width=0.9\textwidth]{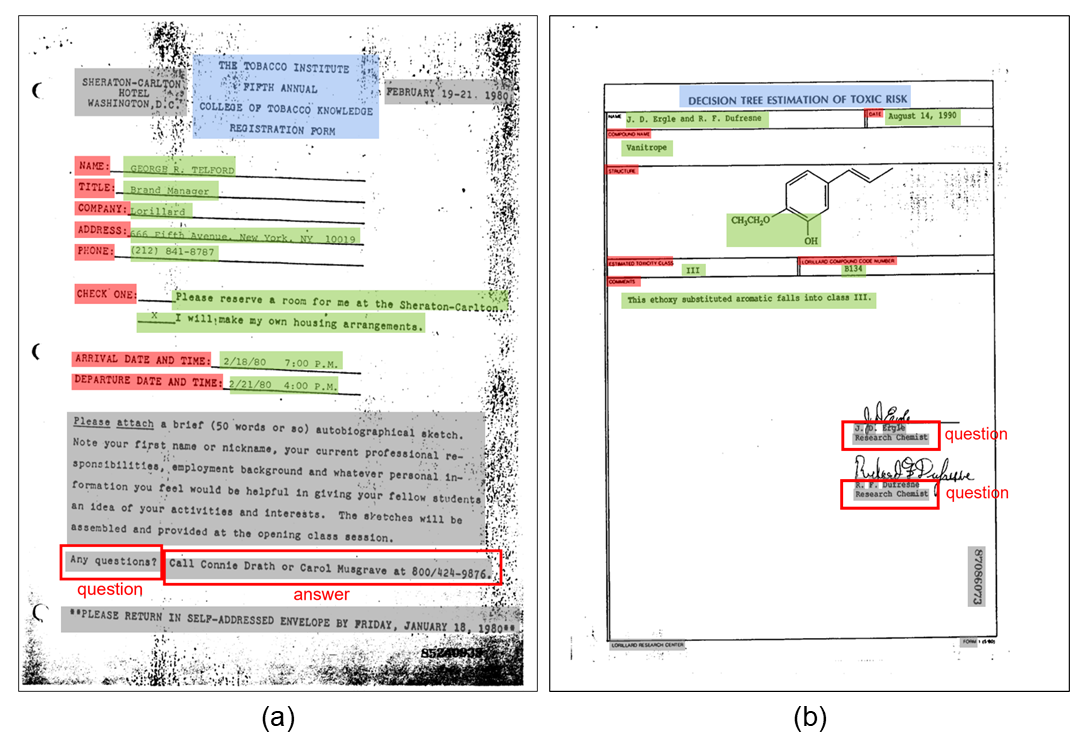}
\caption{
 Visualization of two cases on FUNSD, which are predicted by GPT-4. Blue: ``header", Red: ``question", Grean: ``answer", Grey: ``other". The entities contained within the red box are predicted inaccurately.  
} 
\label{funsdcase1}
\end{figure*}

\begin{figure*}[t]
\centering
\includegraphics[width=0.85\textwidth]{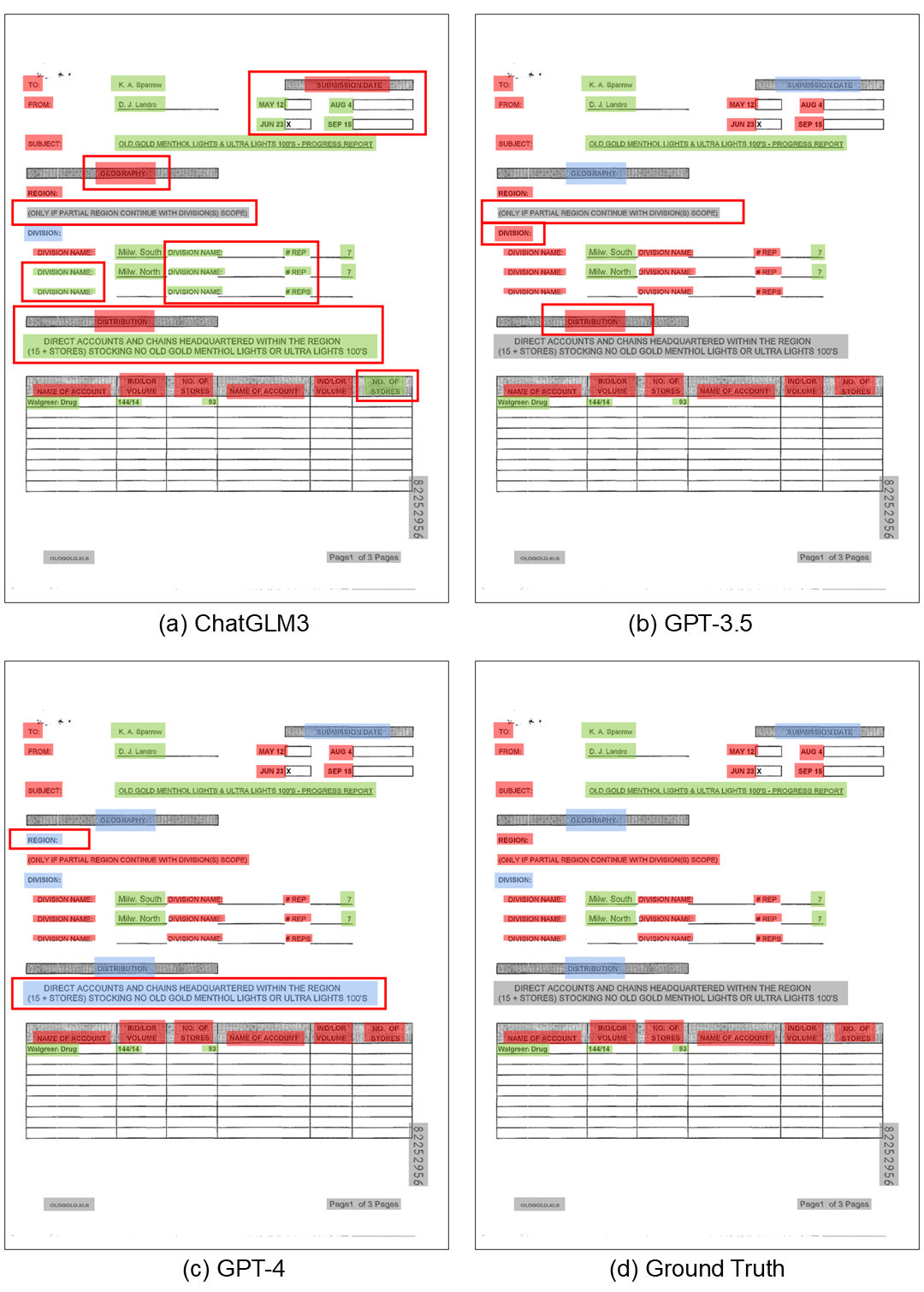}
\caption{
 Visualization of four cases on FUNSD, which are predicted by (a) ChatGLM3, (b) GPT-3.5, (c) GPT-4, and (d) Ground Truth.
 Blue: ``header", Red: ``question", Grean: ``answer", Grey: ``other".  The entities contained within the red box are predicted inaccurately. 
} 
\label{funsdcase2}
\end{figure*}

\begin{figure*}[t]
\centering
\includegraphics[width=0.86\textwidth]{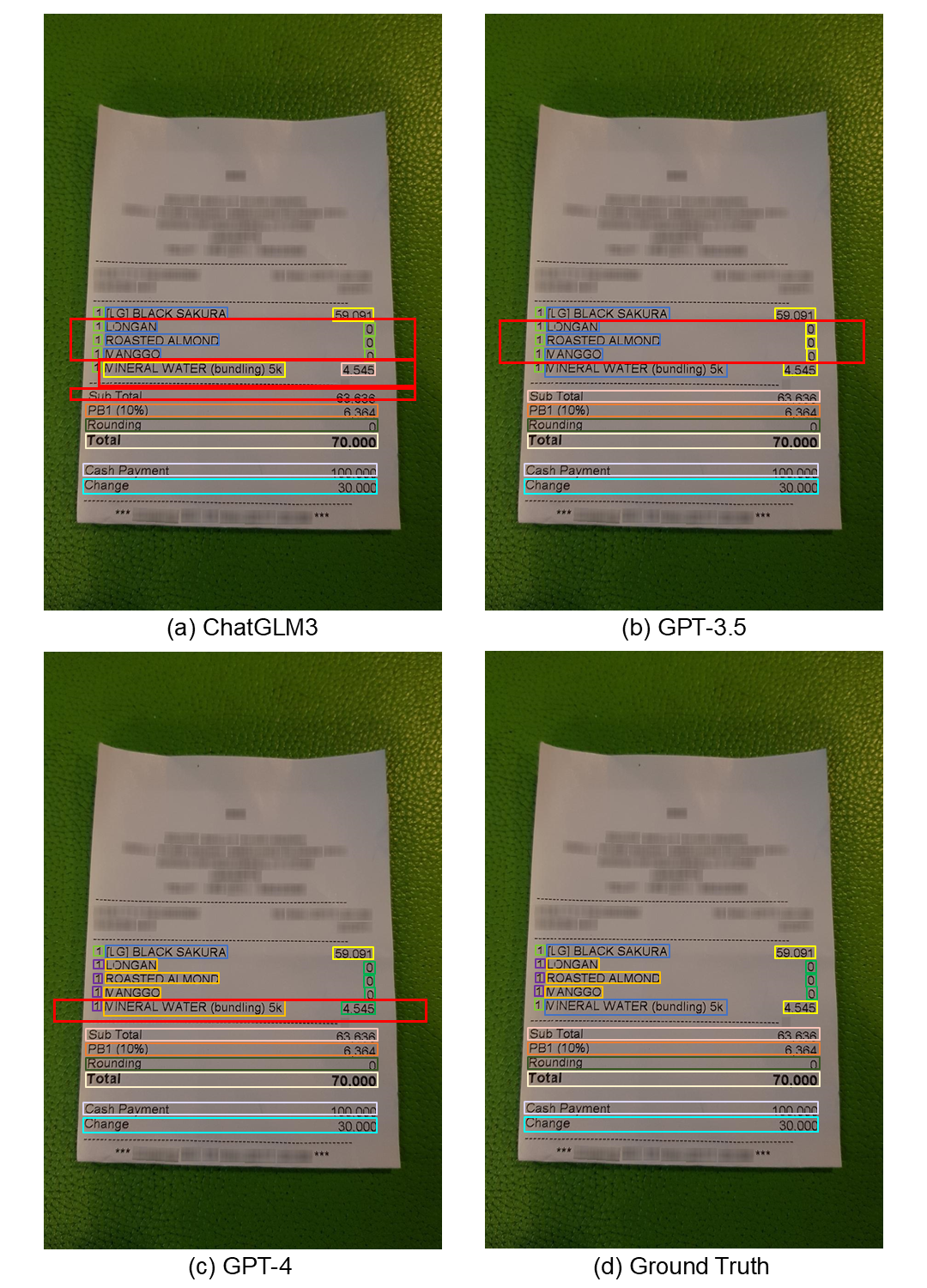}
\caption{
 Visualization of four cases on CORD, which are predicted by (a) ChatGLM3, (b) GPT-3.5, (c) GPT-4, and (d) Ground Truth.
The entities contained within the red box are predicted inaccurately. The corresponding prediction is shown in Figure \ref{cordpredresult}.
} 
\label{cordcase2}
\end{figure*}

\begin{figure*}[t]
\centering
\includegraphics[width=0.9\textwidth]{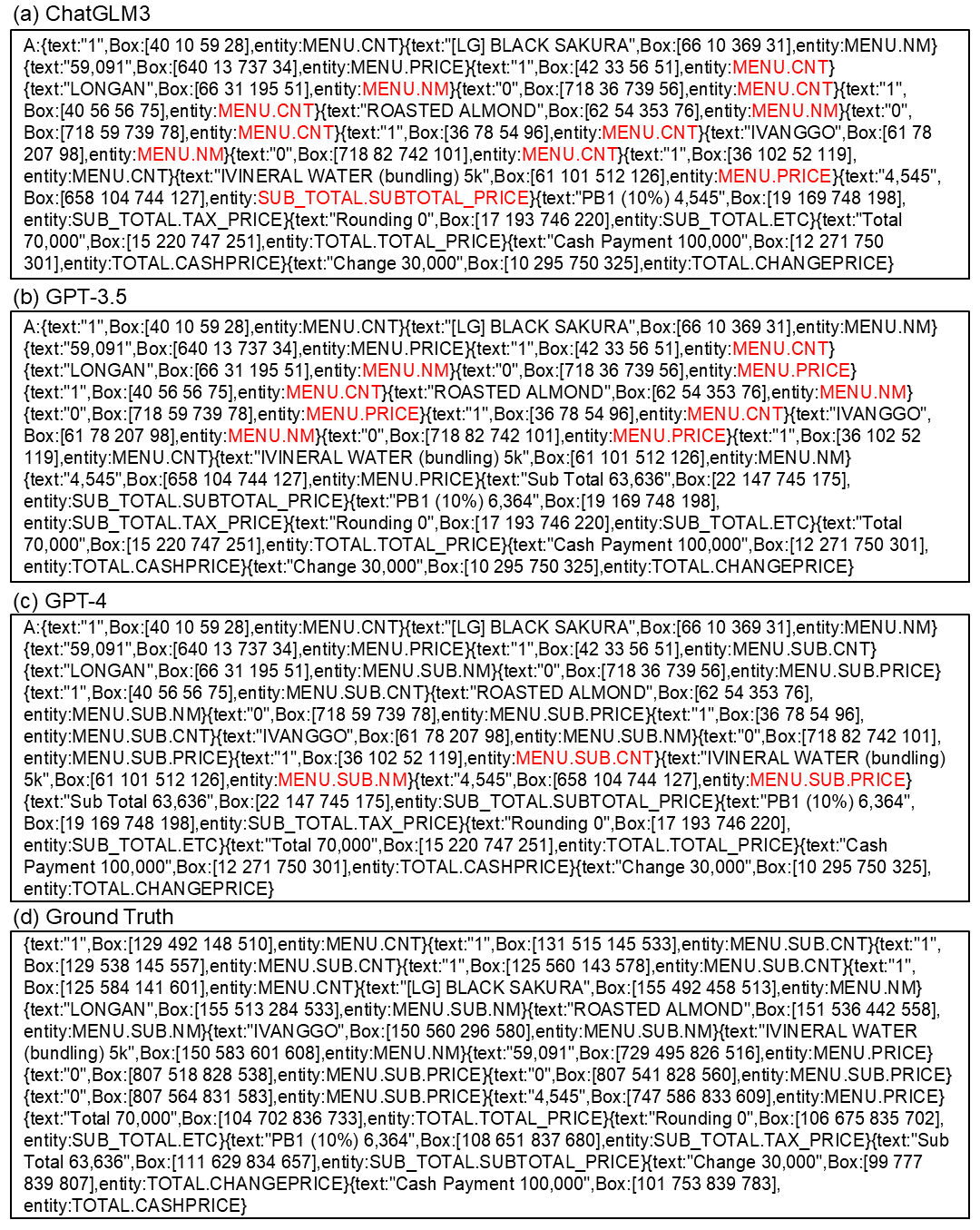}
\caption{
 Examples of prediction by (a) ChatGLM3, (b) GPT-3.5, (c) GPT-4, and (d) Ground Truth on CORD.
The labels of the red color are inaccurate.   
} 
\label{cordpredresult}
\end{figure*}

\begin{figure*}[t]
\centering
\includegraphics[width=\textwidth]{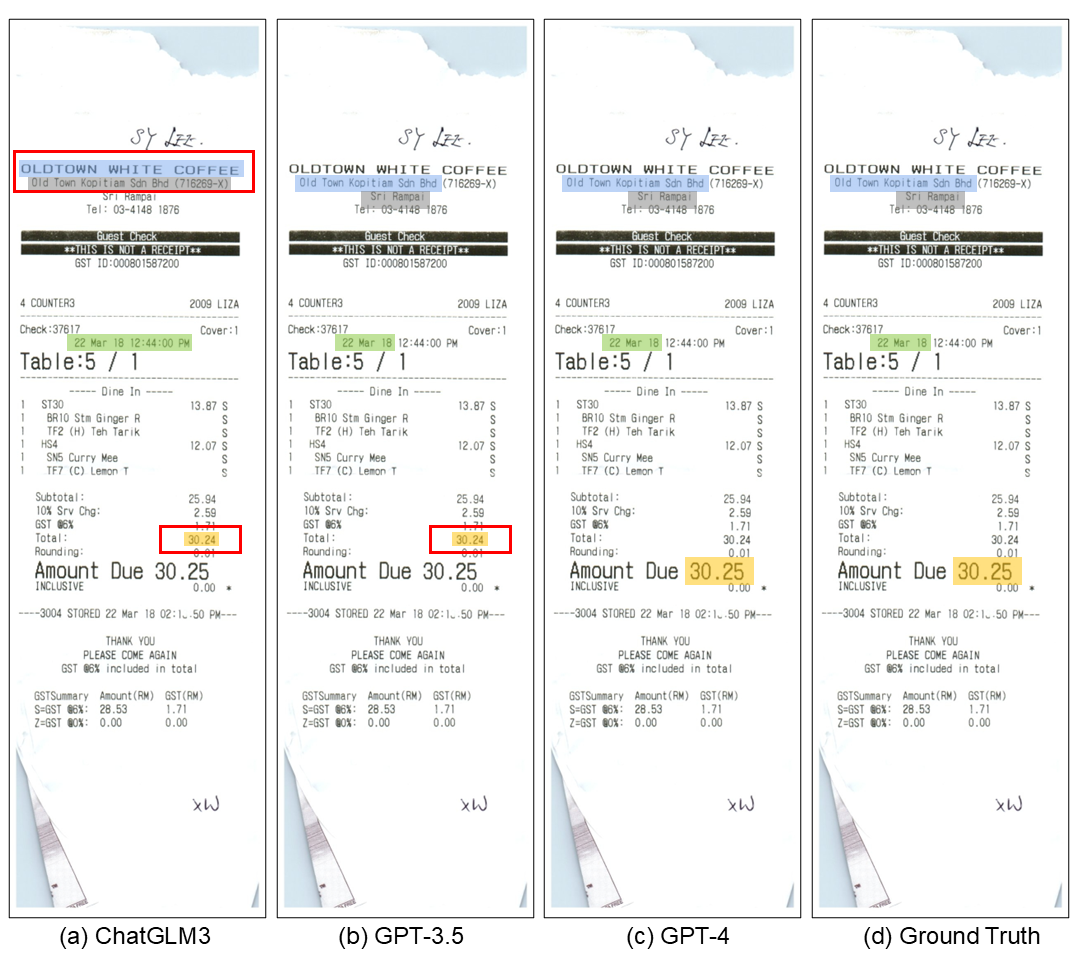}
\caption{
Visualization of four cases on SROIE, which are predicted by (a) ChatGLM3, (b) GPT-3.5, (c) GPT-4, and (d) Ground Truth.
Blue: ``company", grey: ``address", green: ``date", orange: ``total".
The entities contained within the red box are predicted inaccurately. The corresponding prediction is shown in Figure \ref{sroiepredresult}.
} 
\label{sroiecase2}
\end{figure*}

\begin{figure*}[t]
\centering
\includegraphics[width=0.95\textwidth]{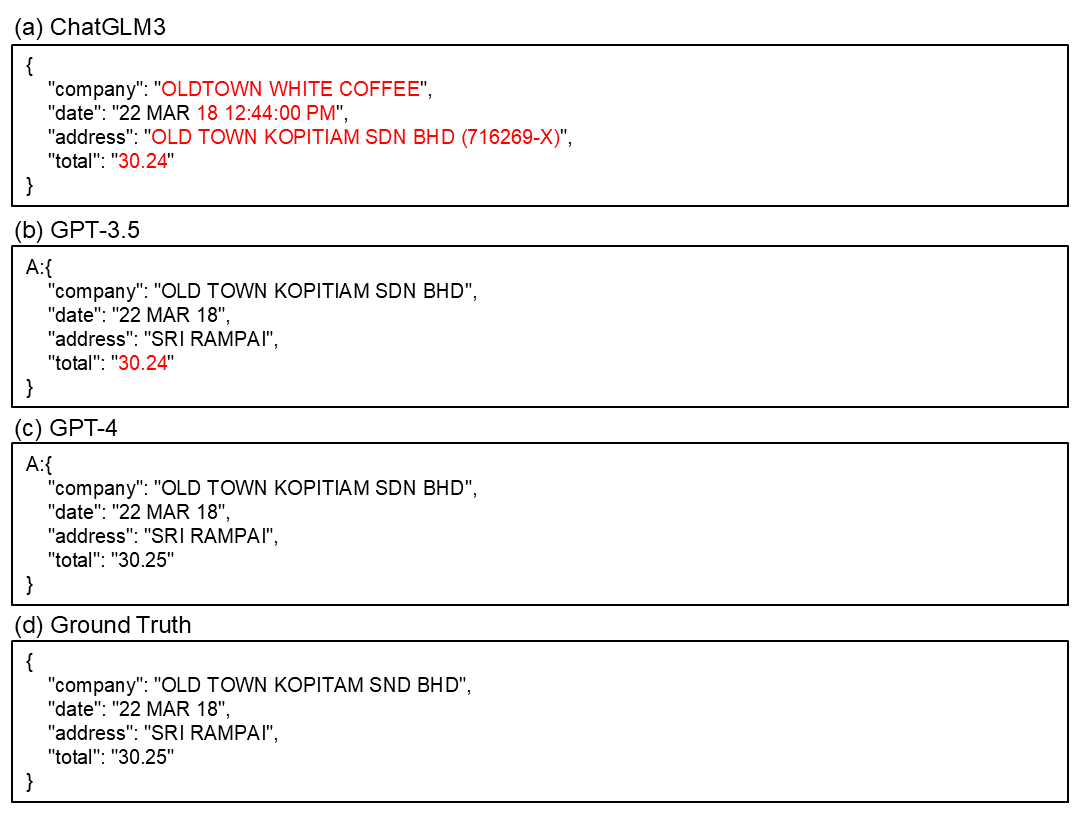}
\caption{
 Examples of prediction by (a) ChatGLM3, (b) GPT-3.5, (c) GPT-4, and (d) Ground Truth on SROIE.
The labels of the red color are inaccurate.   
} 
\label{sroiepredresult}
\end{figure*}

\begin{figure*}[t]
\centering
\includegraphics[width=0.9\textwidth]{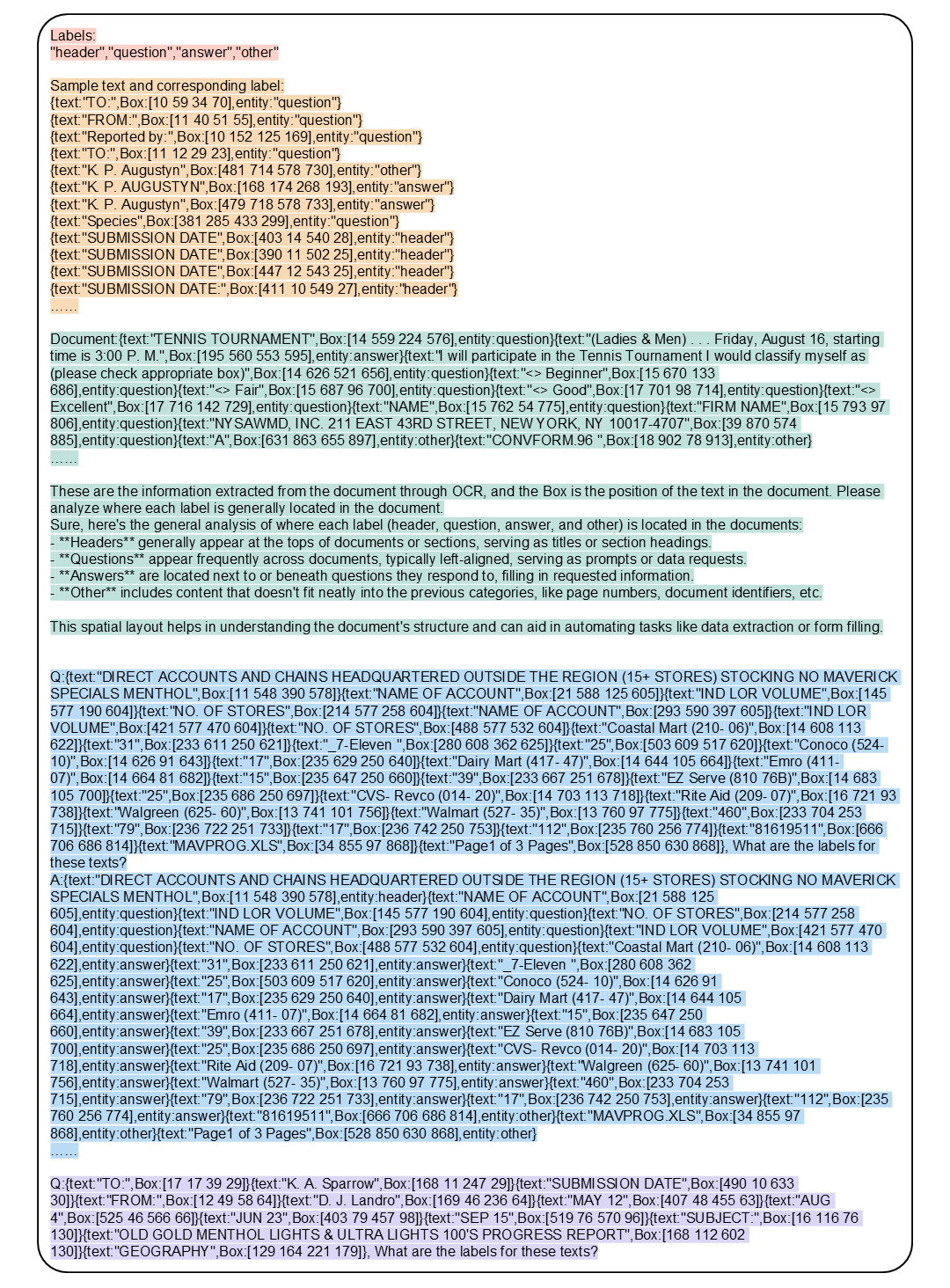}
\caption{
An example of \method prompt for FUNSD dataset. Pink: candidate labels illustration ($C_{\rm cl}$), orange: entity-level text demonstrations ($C_{\rm et}$), green: layout demonstrations ($C_{\rm l}$), blue: entity-level text demonstrations ($C_{\rm dt}$), purple: test question.
} 
\label{funsdprompt}
\end{figure*}

\begin{figure*}[t]
\centering
\includegraphics[width=0.9\textwidth]{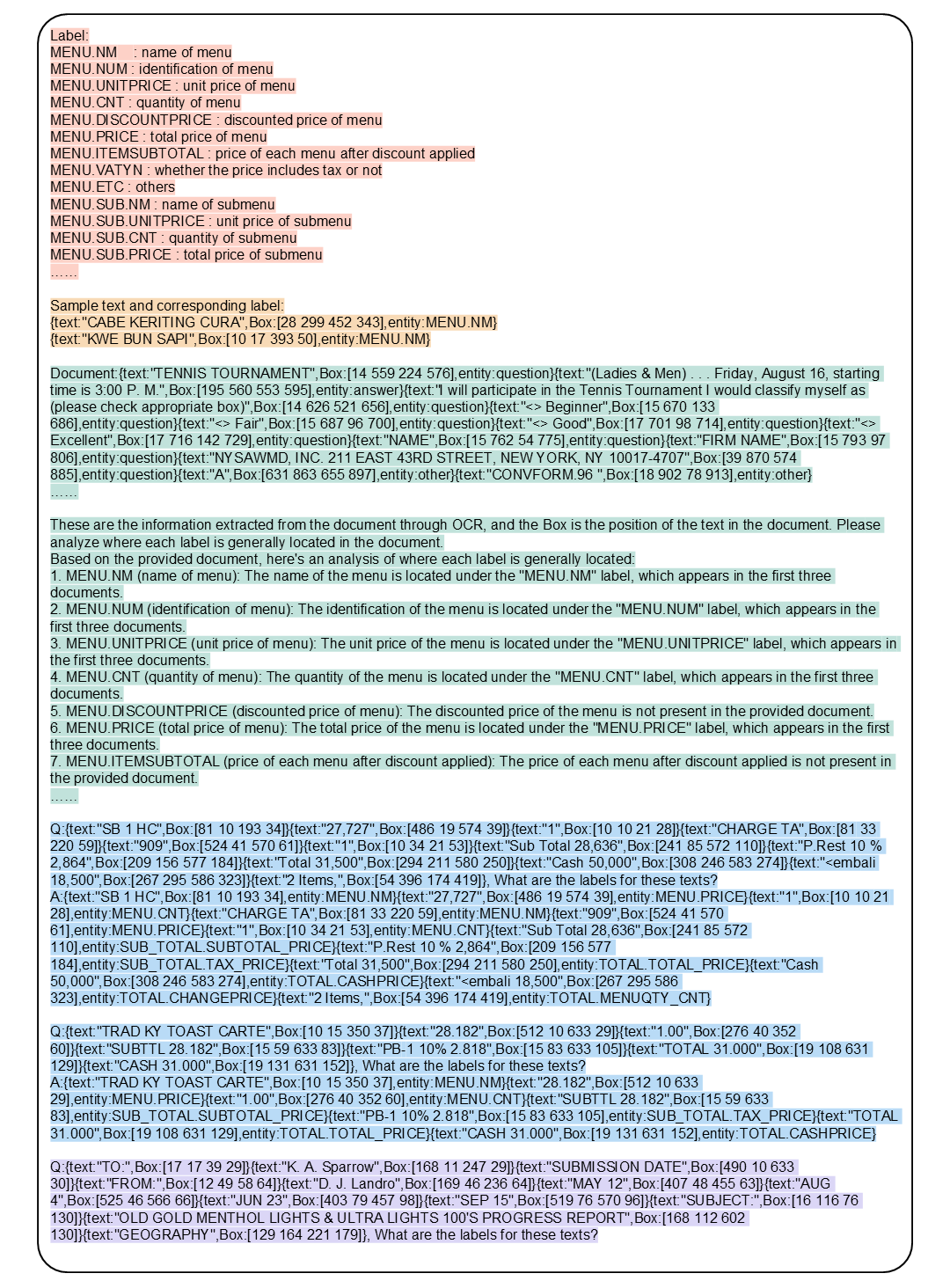}
\caption{
An example of \method prompt for CORD dataset. Pink: candidate labels illustration ($C_{\rm cl}$), orange: entity-level text demonstrations ($C_{\rm et}$), green: layout demonstrations ($C_{\rm l}$), blue: entity-level text demonstrations ($C_{\rm dt}$), purple: test question. 
} 
\label{cordprompt}
\end{figure*}

\begin{figure*}[t]
\centering
\includegraphics[width=0.89\textwidth]{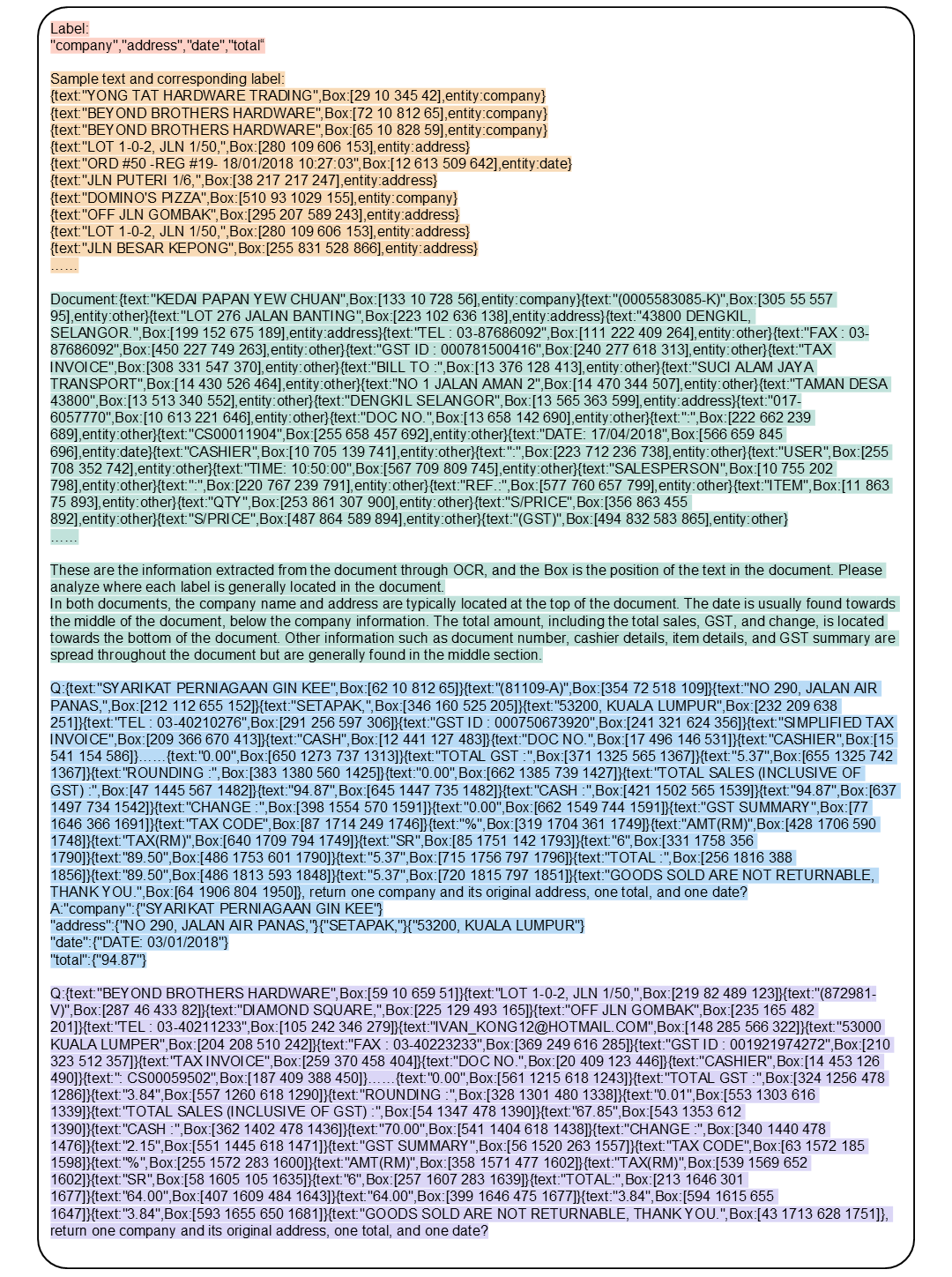}
\caption{
An example of \method prompt for SROIE dataset. Pink: candidate labels illustration ($C_{\rm cl}$), orange: entity-level text demonstrations ($C_{\rm et}$), green: layout demonstrations ($C_{\rm l}$), blue: entity-level text demonstrations ($C_{\rm dt}$), purple: test question. 
} 
\label{sroieprompt}
\end{figure*}

More visual examples of our SAIL are shown in Figure \ref{layoutimagecase} to Figure \ref{sroieprompt}.

\section{Limitations}\label{supp:sec:limit}

Our \method mainly suffers from two limitations. 
\begin{itemize}
    \item 
    The search process incurs an additional time cost, accounting for 13.3\% when using GPT-4.
    \item
    The use of diverse examples increases the total token count.
\end{itemize}
Thus, how to improve the search method and reduce the token count will be the focus of our future work.

\end{document}